\documentclass[10pt,twocolumn,a4,compsoc]{IEEEtran}

\usepackage{times}
\usepackage{epsfig}
\usepackage{graphicx}
\usepackage{amsmath}
\usepackage{amssymb}
\usepackage{subfigure}
\usepackage{multirow}
\usepackage[hyphens]{url}
\usepackage{hyperref}

\usepackage{xcolor}
\usepackage[printwatermark]{xwatermark}
\newwatermark*[allpages,color=red,angle=0,scale=1,xpos=0,ypos=380pt,fontsize=6pt]{
The content of this paper was published in IJCB, 2017. This ArXiv version is from before the peer review. Please, cite the following paper:
\\Ziga Emersic, Dejan Stepec, Vitomir Struc, Peter Peer, Anjith George, Adil Ahmad, Elshibani Omar, Terrance E. Boult, \\Reza Safdari, Yuxiang Zhou, Stefanos Zafeiriou, Dogucan Yaman, Fevziye I. Eyiokur, Hazim Kemal Ekenel: \\“The unconstrained ear recognition challenge”, International Joint Conference on Biometrics, IEEE, 2017
}

\newcommand\blfootnote[1]{%
  \begingroup
  \renewcommand\thefootnote{}\footnote{#1}%
  \addtocounter{footnote}{-1}%
  \endgroup
}

\begin{document}

\date{}
\title{The Unconstrained Ear Recognition Challenge}

\author{\small \v{Z}iga Emer\v{s}i\v{c}, Dejan \v{S}tepec, Vitomir \v{S}truc, Peter Peer\\
\small University of Ljubljana\\
\small Ljubljana, Slovenia\\
{\tt\small ziga.emersic@fri.uni-lj.si}
\\\ \\
\and 
\small Anjith George\\
\small IIT Kharagpur\\
\small Kharagpur, India\\
\ \\
\and
\small Adil Ahmad, Elshibani Omar, Terrance E. Boult\\
\small University of Colorado Colorado Springs\\
\small Colorado Springs, CO, USA\\
\ \\
\and
\small Reza Safdari\\
\small Islamic Azad University\\
\small Qazvin, Iran\\
\ \\
\and
\small Yuxiang Zhou, Stefanos Zafeiriou\\
\small Imperial College London\\
\small London, UK\\
\ \\
\and
\small Dogucan Yaman, Fevziye I. Eyiokur, Hazim K. Ekenel\\
\small ITU Department of Computer Engineering\\
\small Istanbul, Turkey\vspace{-13mm}\\
\ \\ \ \\
}

\maketitle
\thispagestyle{empty}

\begin{abstract}
In this paper we present the results of the Unconstrained Ear Recognition Challenge (UERC), a group benchmarking effort centered around the problem of person recognition from ear images captured in uncontrolled conditions. The goal of the challenge was to assess the performance of existing ear recognition techniques on a challenging large-scale dataset and identify open problems that need to be addressed in the future.  
Five groups from three continents participated in the challenge and contributed six ear recognition techniques for the evaluation, while multiple baselines were made available for the challenge by the UERC organizers. A comprehensive analysis was conducted with all participating approaches addressing essential research questions pertaining to the sensitivity of the technology to head rotation, flipping, gallery size, large-scale recognition and others. The top performer of the UERC was found to ensure robust performance on a smaller part of the dataset (with 180 subjects) regardless of image characteristics, but still exhibited a significant performance drop when the entire dataset comprising $3,704$ subjects was used for testing. \vspace{-4mm}  
\end{abstract}

\blfootnote{The contents of this paper presented at the International Joint Conference on Biometrics 2017, \url{http://www.ijcb2017.org}.}

\section{Introduction}\label{Sec: Introduction}
Recognizing people from ear images with automatic machine-learning techniques represents a challenging problem that is of interest to numerous application domains. Past research in this area has mostly been focused on images captured in controlled conditions and near perfect recognition performance has already been reported on many of the available (laboratory-like) ear datasets, e.g.,~\cite{chan2012reliable,fabate2006ear,guo2008ear,kumar2013robust,Pflug2014a}. The literature on unconstrained ear recognition, on the other hand, is relatively modest and the performance of existing ear recognition techniques on so-called data captured \textit{in the wild} is not well explored. It is not completely clear how existing techniques are able to cope with ear-image variability encountered in unconstrained settings and how the recognition performance is affected by factors such as head rotation, image resolution, occlusion or gallery size.     
\begin{figure}[tb] \begin{center}
\includegraphics[width=0.99\columnwidth]{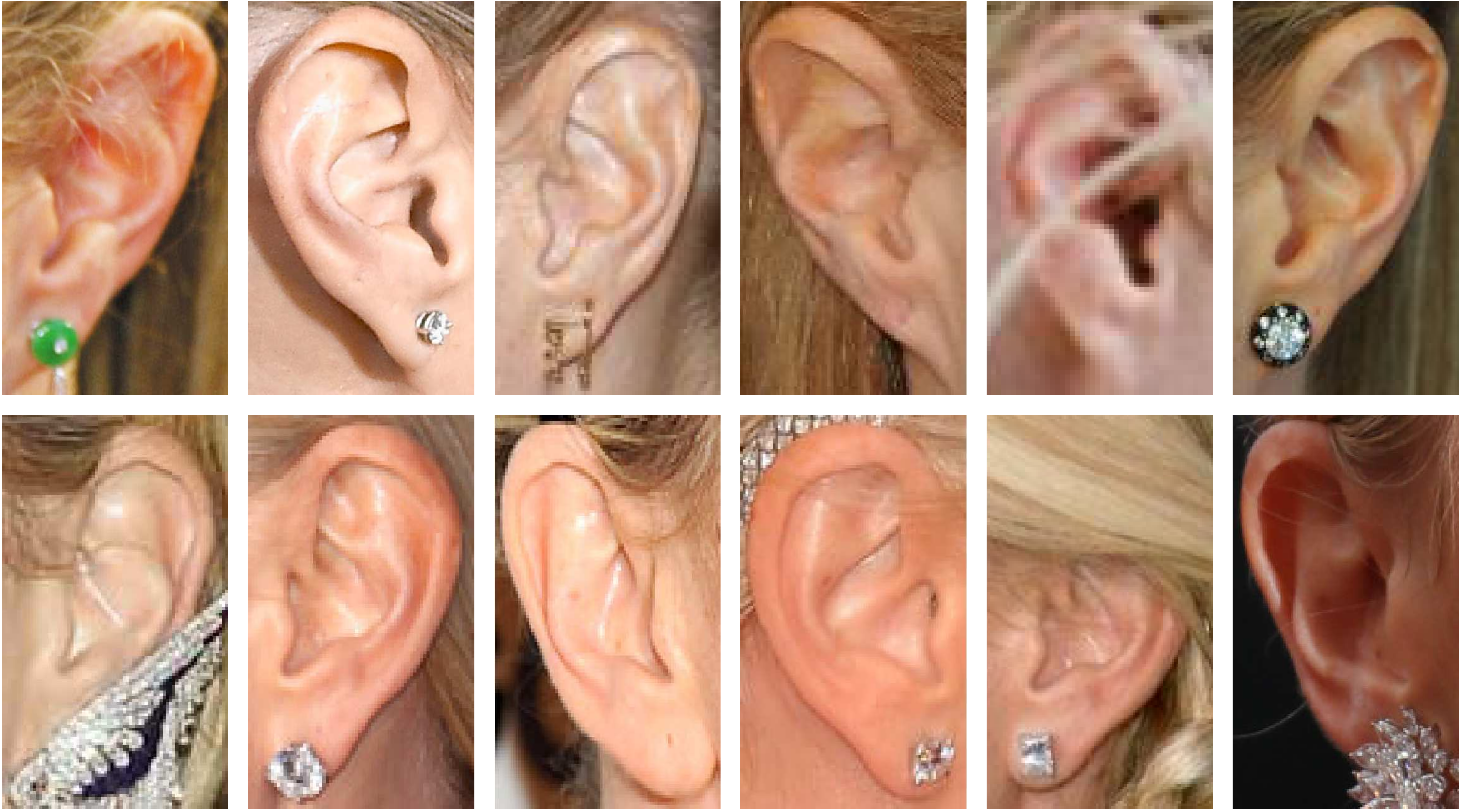}
\caption{Sample images from the UERC dataset. The images are tightly cropped and belong to left and right ears. The variability of the dataset is across ear orientation, occlusion, gender, ethnicity and resolution. Images in one row belong to the same subject.\vspace{-7mm}}
\label{Fig: AWE_crop_jpg}
\end{center}\end{figure}

To address this gap and study the problem of ear recognition from unconstrained ear images, the Unconstrained Ear Recognition Challenge (UERC), a first of its kind group benchmarking effort, was organized in the scope of the $3$rd International Joint Conference on Biometrics (IJCB $2017$). The goal of the challenge was to exploit the combined expertise of multiple research groups for a comprehensive evaluation of ear recognition technology, to consolidate research and identify open challenges through experiments on a common dataset and a well defined experimental protocol. As illustrated in Fig.~\ref{Fig: AWE_crop_jpg}, a large and challenging dataset of ear images captured in unconstrained settings was collected specifically for the UERC and made available to the participants for algorithm development and testing. Five groups (from the US, UK, Turkey, Iran and India) took part in the challenge and submitted results for a total of six recognition approaches. A detailed analysis was then conducted investigating various aspects of the submitted techniques, such as sensitivity to head rotation (separately for pitch, roll and yaw angles), impact of gallery size and the ability of the techniques to scale with larger probe and gallery sets.      

The research and development efforts of the UERC organizers and participating research groups resulted in the following contributions that are presented in this paper:
\begin{itemize}
\item The first group benchmarking effort of ear recognition technology on a large datasets of unconstrained ear images corresponding to several thousands of subjects.\vspace{-2mm}
\item A new and challenging dataset of ear images collected from the internet, which (with a total of $11,804$ images of $3,706$ identities) is among the largest ear datasets publicly available to the research community.\vspace{-2mm}
\item An analysis of the characteristics of different ear recognition techniques and identification of the main factors affecting performance.
\end{itemize}


\section{Related work}\label{Sec: Related}

Challenges and group evaluations have a long history in the field of computer vision and biometrics in particular. The goal of these events is to benchmark existing work in a certain problem domain, provide a snapshot of the current state-of-technology and point to open issues that need to be addressed in the future. As a result of 
these challenges, significant advancements have been made over the years that pushed the capabilities of 
computer-vision technology.  

Examples of recent challenges that had a profound impact on the field of computer vision are the ImageNet Large Scale Visual Recognition Challenges (ILSVRC)~\cite{deng2009imagenet,ILSVRC15}, which focus on image classification and object localization problems, the Visual Object Tracking (VOT)~\cite{Kristan2016a,kristan2015visual,kristan2016novel} challenges that aim at evaluating various solutions to object tracking in videos, and the ChaLearn Looking at People~\cite{escalerachalearn,escalera2016chalearn,wan2016chalearn} series of challenges, where human-centric vision problems are at the center of attention.

Among past biometrics-oriented challenges, events focusing on fingerprints~\cite{maio2004fvc2004,maio2002fvc2000,maio2002fvc2002} and facial images~\cite{beveridge2014ijcb,huang2007labeled,phillips2005overview,scheirer2016report} have likely been the most visible. Recent years have seen challenges and competitions on other modalities and biometric sub-problems ranging from iris~\cite{bowyer2012results}, speaker~\cite{greenberg2014nist,marcel2010results}, sclera~\cite{das2016ssrbc}, finger-vein~\cite{xian2015icb,ye2016fvrc2016}, or keystroke dynamics~\cite{morales2016keystroke} recognition to spoof detection~\cite{chakka2011competition,chingovska20132nd,tome20151st}, liveness detection~\cite{ghiani2013livdet,marcialis2009first,yambay2014livdet}, segmentation~\cite{das2016ssrbc,dasa2015ssbc} and others. 

Here, we add to the outlined body of work and present the results of the (first of its kind) ear recognition challenge. The challenge focuses on the under-explored problem of ear recognition from images captured in unconstrained environments and aims at presenting a comprehensive analysis of the existing ear recognition technology and at providing the community with a new dataset for research in this area. 

\section{Methodology}\label{Sec: methods}

In this section we describe the methodology used for the UERC. We first introduce the experimental dataset and then present the experimental protocol and performance metrics used for the evaluation.
\begin{table*}[!tb]
\renewcommand{\arraystretch}{1.15}
\caption{Overview of the UERC data partitioning. The annotated part of the UERC data, i.e., the AWE dataset, was split between the training and testing parts. The partitioning was disjoint in terms of subjects. \vspace{1mm}}
\label{Tab: experimental protocol}
\centering
\small
\setlength\tabcolsep{5.5pt}
\begin{tabular}{l lcccc}
 \hline  \hline 
UERC data partition 				& 	Image origin 			& \#  Images 	& \# Subjects 	& \# Images per subject 	& Total \# Images (\# Subjects)\\
 \hline 
\multirow{ 2}{*}{Training part}		&   Auxiliary AWE dataset	& $804$ 	& $16$			&	Variable			&  \multirow{ 2}{*}{$2,304 \text{ }(166)$}\\
									& 	Part of AWE dataset		& $1,500$    & $150$			& 	$10$				& \\
\hline                                     
\multirow{ 2}{*}{Testing part}		& 	Part of AWE dataset		& $1,800$ 	& $180$			& 	$10$				& 	\multirow{ 2}{*}{$9,500 \text{ }(3,540)$}\\
									&   Newly collected			& $7,700$ 	& $3,360$		& 	Variable 			& \\
\hline 
\hline
\end{tabular}
\end{table*}

\subsection{The UERC dataset}\label{SSec: dataset}

The data used for the UERC is a blend of two existing and one newly collected dataset and contains a total of  $11,804$ ear images of $3,706$  subjects. The core part of the data comes from the Annotated Web Ears (AWE)~\cite{ZigaSurvey2017} dataset and features $3,300$ ear images of $330$ subjects. Images from this part contain various annotations, such as the extent of head rotation (in terms of pitch, yaw and roll angles), gender, level of occlusion, ethnicity and others. A detailed description of the AWE dataset is available from~\cite{ZigaSurvey2017}. The second part of the UERC data, i.e., $804$ images of $16$ subjects, comes from the  auxiliary AWE dataset\footnote{Available from \url{http://awe.fri.uni-lj.si/download}.}, while the majority of images, i.e., $9,500$ images of $3,540$ subjects, was gathered exclusively for the UERC.

The data-collection procedure for the newly gathered images was semi-automatic. Similarly to other datasets gathered   \textit{in the wild}~\cite{huang2007labeled, kemelmacher2016megaface}, a list of celebrities was first generated and web crawlers were used to pull candidate face images from the internet. An automatic ear segmentation procedure based on convolutional encoder-decoder networks~\cite{ZigaEarDetection2017} was then applied to the candidate images to identify potential ear regions. Finally, the segmented ears were manually inspected and miss detections and partial segmentations were discarded. The final images included in the UERC dataset were tightly cropped, but  were not normalized to a common side. Thus, original images of left and right ears are featured in the dataset. A major characteristic of the UERC images is the large variability in size, since the smallest images containing only a few hundred pixels, whereas the largest contain close to $400k$ pixels. The average pixel count per image is $3,682$. 

Some examples of the UERC ear images are shown in Fig.~\ref{Fig: AWE_crop_jpg}. As can be seen, the images were not captured in controlled (laboratory-like) environments, the variability of the images is, therefore, substantial. The UERC dataset represents, to the best of our knowledge, the first attempt to gather data for ear-recognition research from the web, and as shown in~\cite{ZigaSurvey2017} is significantly more challenging than competing ear datasets where performance is mostly saturated. 

\subsection{Protocol and performance metrics}\label{SSec: protocol}

For the evaluation, the UERC ear images were split into two disjoint groups, one for training and one for testing. The training part contained images from the auxiliary AWE dataset and around half of the original AWE data, whereas the testing part comprised the other half of the AWE data and all newly collected images. A summary of the data split is given in Table~\ref{Tab: experimental protocol}.

The training part of the UERC data was used to train or fine-tune potential models (e.g., deep models, classifiers, etc.), while the testing part was used exclusively for the  performance evaluation. Using any images from the testing part for training was not allowed. The UERC participants were asked to submit a similarity matrix of size $7,742\times 9,500$ to the organizers, which served as the basis for the performance assessment. The similarity matrix was produced by matching the $7,742$ probe images (of $1,482$ subjects) to all $9,500$ gallery images (belonging to $3540$ subjects).  The gallery featured all images from the testing part of the UERC data, while the $7,742$ probe images represented a subset of these $9,500$ galleries from subjects having  at least $2$ images in the dataset. Each similarity score had to be generated based solely on the comparison of two ear images and no information about other subjects in the testing part of the UERC data was allowed to be used for the score generation. A sample script that implemented the protocol was distributed among the UERC participants to ensure that all submitted similarity matrices were generated consistently with the same training and testing data. 

The performance of the submitted approaches was measured through recognition (identification) experiments and Cumulative Match Score (CMC) curves were used to visualize the results. Additionally, performance was measured with the following quantitative metrics: \textit{i)} the recognition rate at rank one (rank-$1$), which corresponds to the percentage of probe images, for which an image of the correct identity was retrieved from the gallery as the top match, \textit{ii)} the recognition rate at rank five (rank-$5$), which corresponds to the percentage of probe images, for which an image of the correct identity was among the top five matches retrieved from the gallery, and \textit{iii)} the area under the CMC curve (AUC), which analogous to the more widely used Receiver Operating Characteristic (ROC) AUC, measured the overall performance of the recognition approached at different ranks. The latter was computed by normalizing the maximum  rank (i.e., the number of distinct gallery identities) of the experiments to one. 

To study the performance of the submitted approaches and analyze their characteristics, specific parts corresponding to different data labels (e.g., probe images with specific yaw, roll or pitch angles) were sampled from the  similarity matrix and evaluated. 

\section{Participating approaches}\label{Sec: Participants}

In this section we describe all participating approaches of the UERC. Five groups submitted results 
and several baselines were made available by the challenge organizers. 

\subsection{Baseline approaches}

Eight baseline approaches were provided for the UERC by the organizers from the University of Ljubljana with the goal of ensuring references implementations of the experimental protocol and an initial estimate of the difficulty of the dataset. Two of these baselines were then selected for most of the experimental analysis in Section~\ref{Sec: exper} to keep the results uncluttered. The first  was a descriptor-based technique exploiting local binary patterns~\cite{pietikainen2011local} (LBP-baseline) and the second a deep learning approach based on the 16-layer VGG network architecture~\cite{parkhi2015deep} (VGG-baseline). 
The baselines were implemented within the UERC toolkit 
and distributed among the participants as a starting point for their research work. A detailed description of the two selected baselines is given below.  

\textbf{LBP-baseline:} The first UERC baseline included in the results section follows the usual pipeline used to compute image descriptors based on local binary patterns (LBPs)~\cite{pietikainen2011local}. Patches of size $16\times 16$ pixels are first sampled from the images using a sliding window approach and a step size of $4$ pixels. Each patch is then encoded with uniform LBPs computed with a radius of $R=2$ and a local neighborhood size of $P=8$. $59$-dimensional histograms are calculated for each patch and the histograms of all image patches are concatenated to form the final ear descriptor. The similarity of two ear images is measured with the cosine similarity between the corresponding LBP descriptors.   

\textbf{VGG-baseline:} The second selected UERC baseline is built around a deep learning model, specifically, around a convolution neural network (CNN) based on the $16$-layer VGG architecture from~\cite{simonyan2014very}. The model consists of multiple convolutional layers which, different from competing models, e.g.,~\cite{krizhevsky2012imagenet}, use small filters (of size $3\time 3$ pixels) to reduce the number of parameters that need to be learned during training. The convolutional layers are interspersed with max-pooling layers that reduce the size of the intermediate image representations and followed by two fully-connected layer. The output of second fully-connected layer is used as a $4,096$-dimensioanl ear descriptor. The VGG-baseline model is trained from scratch using the training data from the UERC dataset and aggressive data augmentation. A detailed description of the training procedure is presented in~\cite{ziga@bwild}. Two ear images are matched by computing the cosine distance between the corresponding image descriptors.

\subsection{University of Colorado Colorado Springs}

The group from the University of Colorado Colorado Springs (UCCS) approached the challenge with a novel ear descriptor based on \textit{Chainlets} and utilized a recent deep-learning-based approach for contour detection~\cite{deep_contour_2015} to facilitate the descriptor computation procedure. 

\begin{figure}[tb] \begin{center}
\includegraphics[width=0.99\columnwidth]{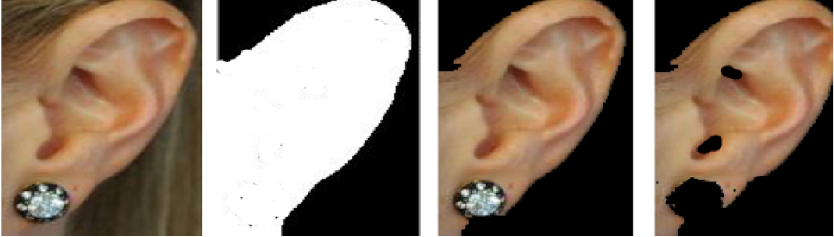}
\caption{Illustration of the UCCS preprocessing procedure (from left to right): the input image, the computed binary ear mask, the masked input image, the input image without accessories and occlusions. The preprocessing aims at removing non-ear regions that could adversely affect the descriptor computation process.\vspace{-6mm}}
\label{UCCS_preproc}
\end{center}\end{figure}

The UCCS approach starts with a preprocessing procedure (illustrated in Fig.~\ref{UCCS_preproc}) aimed at detecting the ear region in the image. With the preprocessing procedure  the input image is first converted to gray-scale and resized to $100\times 100$ pixels. Next, contrast limited adaptive histogram equalization (CLAHE~\cite{zuiderveld1994contrast}) is applied for contrast enhancement and a binary version of the image is produced through intensity thresholding. Morphological operations such as dilation and opening  are  employed to remove noise and to help accentuate the structural information of the ear. The processed binary image is then analyzed and the largest connected region is selected as the ear mask and used to exclude all background pixels from the image that could adversely affect the descriptor computation procedure. To also remove potential earrings and other accessories left in the image, color segmentation focusing on skin-tone values in the HSV color model is used. The result of this procedure is a clean region-of-interest as shown in the right most image of Fig.~\ref{UCCS_preproc}. 
\begin{table*}[!tb]
\renewcommand{\arraystretch}{1.1}
\caption{Comparative overview of the participating approaches. The table provides a short description of each approach, information on whether ear alignment and flipping was performed and the model size (if any). The model size is approximate and given in MB. \vspace{1mm}}
\label{Tab: method comparison}
\centering
\footnotesize
\resizebox{\textwidth}{!}{%
\begin{tabular}{l lcccc}
 \hline  \hline 
Approach & Description & Descriptor type & Ear alignment & Flipping & Model size\\
 \hline 
UCCS			&	Descriptor-based (chainlets)	  								& 	Hand-crafted			& 	No	& No  & no model used\\
IAU				& 	VGG network (trained on ImageNet) and transfer learning 		&   Learned					&	No	& No  & $\sim 100$ MB\\
ICL				&	Deformable model and Inception-ResNet 							& 	Learned					&	Yes	& Yes  & $\sim 100$ MB \\
IITK 			&   VGG network (trained on the VGG face dataset) 					& 	Learned 				&   No  & Yes & $\sim 100$ MB \\
ITU I			&   VGG network (trained on ImageNet) and transfer learning			&	Learned					&	No	& Yes & $\sim 100$ MB \\
ITU II 			& 	Ensemble method (VGG-network + LBP)								&	Learned + Hand-crafted	&	No	& Yes & $\sim 100$ MB \\
\hline
LBP-baseline 	&   Descriptor-based (uniform LBPs)									&	Hand-crafted			&	No	& No  & no model used \\
VGG-baseline 	&  	VGG network trained solely on the UERC training data								&  	Learned					&	No	& No  & $\sim 500$ MB \\
\hline 
\hline
\end{tabular}
}
\end{table*}

Once the the input image is segmented and the region-of-interest is detected, the UCCS approach proceeds with the descriptor calculation step and computes a chainlet-based image descriptor from the cleaned ear area. The main idea here is that the appearance and shape of an object can be well described by the density of Relative Chain Codes, which encode rotation-invariant edge directions. 
The descriptor is similar in essence to HOG~\cite{dalal2005histograms} but relies on longer connected edges and provides a richer and rotation invariant description of edge orientation. 

To compute chainlets, image pixels are first grouped into cells, and the direction of each pixel is computed through a Relative Chain Code~\cite{terry2017chainlets}. In the case of ear recognition, cells of size $8 \times 8$  pixels are selected. Also, computation of a Relative Chain Code Histogram (CCH) is done over the code directions within each cell. To ensure invariance to contrast changes, neighboring cells are grouped into ``blocks'' and the CCHs of all cells from a given block are jointly normalized. The normalization is performed for all image blocks in a sliding window (or better said sliding block) manner. The chainlets descriptor is finally formed by concatenating the normalized CCHs from all (overlapping) image blocks. To measure the similarity of two ear images required for the UERC evaluation procedure, the chi-square distance between the corresponding chainlets descriptors is calculated. A detailed description of the UCCS approach is available from~\cite{terry2017chainlets}.

\subsection{Islamic Azad University}
The group from the Islamic Azad University (IAU) participated in the challenge with an ear recognition approach exploiting the $16$-layer VGG network from~\cite{simonyan2014very} and transfer learning. The approach is similar to the UERC VGG-baseline but relies on weights learned from the ImageNet Large Scale Visual Recognition Challenge (ILSVRC-$2014$). The main idea here is to keep part of the pretrained VGG model in tact, while retraining other parts that are relevant for the new problem domain, i.e., ear recognition. 

The amount of  training data available in the UERC dataset is relatively modest and differs from ImageNet both in terms of image content as well the number of classes. It might, hence, not be optimal to train a classifier on top of the network, as image representations generated by the higher network layers are commonly problem- and dataset-specific. A potentially better approach is to rely on activations  from earlier network layers and use these to train a linear classifier for the new problem domain. The IAU approach, therefore, adds two fully-connected layers on top of the $7$th layer of the pretrained VGG model. The pretrained weights of the early layers are frozen and kept unchanged, while the newly added fully-connected layers are trained from scratch with the UERC training data using stochastic gradient descend (SGD) and the softmax loss function. In order to prevent overfitting, a dropout rate of $0.7$ and $L_2$ weight decay regularization are applied on the fully-connected layers~\cite{srivastava2014dropout}. Additionally, data augmentation~\cite{krizhevsky2012imagenet} is used to increase the amount of the data available for training. After convergence, the last fully-connected layer (i.e, the classifier) is removed and the output of the penultimate layer is used an image descriptor.

Once the model is fully trained, it is used as a ``black-box'' feature extractor. A given ear image is simply rescaled to a size of $64\times 64$ pixels, mean centered and passed through the network. The output of the model is a $512-$dimensional ear descriptor that can be matched against other ear descriptors using the cosine similarity measure. 

\subsection{Imperial College London}

The group from the Imperial College London (ICL) participated in the UERC with an approach build around Statistical Deformable Models (SDMs)~\cite{zhou2016estimating} and Inception-ResNets~\cite{szegedy2016inception}. The SDM was used for dense ear alignment and the Inception-ResNet for descriptor computation.

For the ear-alignment model the ICL group used an in-house dataset of $605$ ear images annotated with $55$ landmarks\footnote{\url{https://ibug.doc.ic.ac.uk/resources/ibug-ears/}}. A specially designed SDM, capable of handling training data with inconsistent annotations, was then trained with the annotated ears and used to densely align all images from the UERC dataset. During the alignment step, the ear images were flipped and the deformable model was fitted to the original as well as flipped images. The image that resulted in a lower loss during model fitting was  chosen as the basis for descriptor computation.

In the feature-learning part of the ICL approach, an Inception-ResNet~\cite{szegedy2016inception} was trained from scratch using aligned ear images from the training part of the UERC dataset. The Inception-ResNet architecture was chosen for this part because of its competitive performance and the ability to be trained fast due to the residual connections of the model. The network was trained with a marginal- and softmax-cross-entropy loss for $80$ epochs. After the training, the classification layer was removed and the output of the model was used as a $512$-dimensional ear descriptor.

In the evaluation stage, the probe images were first aligned with the trained SDM and ear descriptors were computed with the Inception-ResNet model. A similarity score for two images was produced based on the $L_2$-norm between the corresponding ear descriptors. 

\subsection{Indian Institute of Technology Kharagpur}

The group from Indian Institute of Technology Kharagpur (IITK) approached the UERC with a two-step technique that first detected whether the input images belong to the left or the right ear and then computed ear descriptors for matching from the side-normalized images using a pretrained $16$-layer VGG model. 

The IITK group trained a simple SVM model over HOG descriptors to classify whether the given image corresponds to the left or the right ear. For this step, the training images were resized to $30 \times 60$ pixels prior to the extraction of the HOG features. The output of the classifier was then used to flip the images to a common reference. The side-normalized images were resized to a fixed size of $224\times 224$ pixels and fed to the pretrained VGG face network~\cite{parkhi2015deep}. Because the last few fully connected layers of the VGG model were tuned specifically for face recognition, only the feature maps from the convolutional layers were considered and the pooled output of these feature maps was used as the ear descriptor. For similarity score calculation, the cosine distance between two ear descriptors was adopted. 

\subsection{Istanbul Technical University}

The participants from the Istanbul Technical University (ITU) submitted two approaches to the UERC. The first was a deep learning approach based on the VGG network~\cite{simonyan2014very} architecture (ITU-I hereafter) and the second an ensemble approach combining the VGG model with hand-crafted LBP descriptors (ITU-II from hereon)~\cite{pietikainen2011local}. A brief summary of both approaches is given below.
\begin{figure}
  \centering
  \includegraphics[width=0.9\columnwidth]{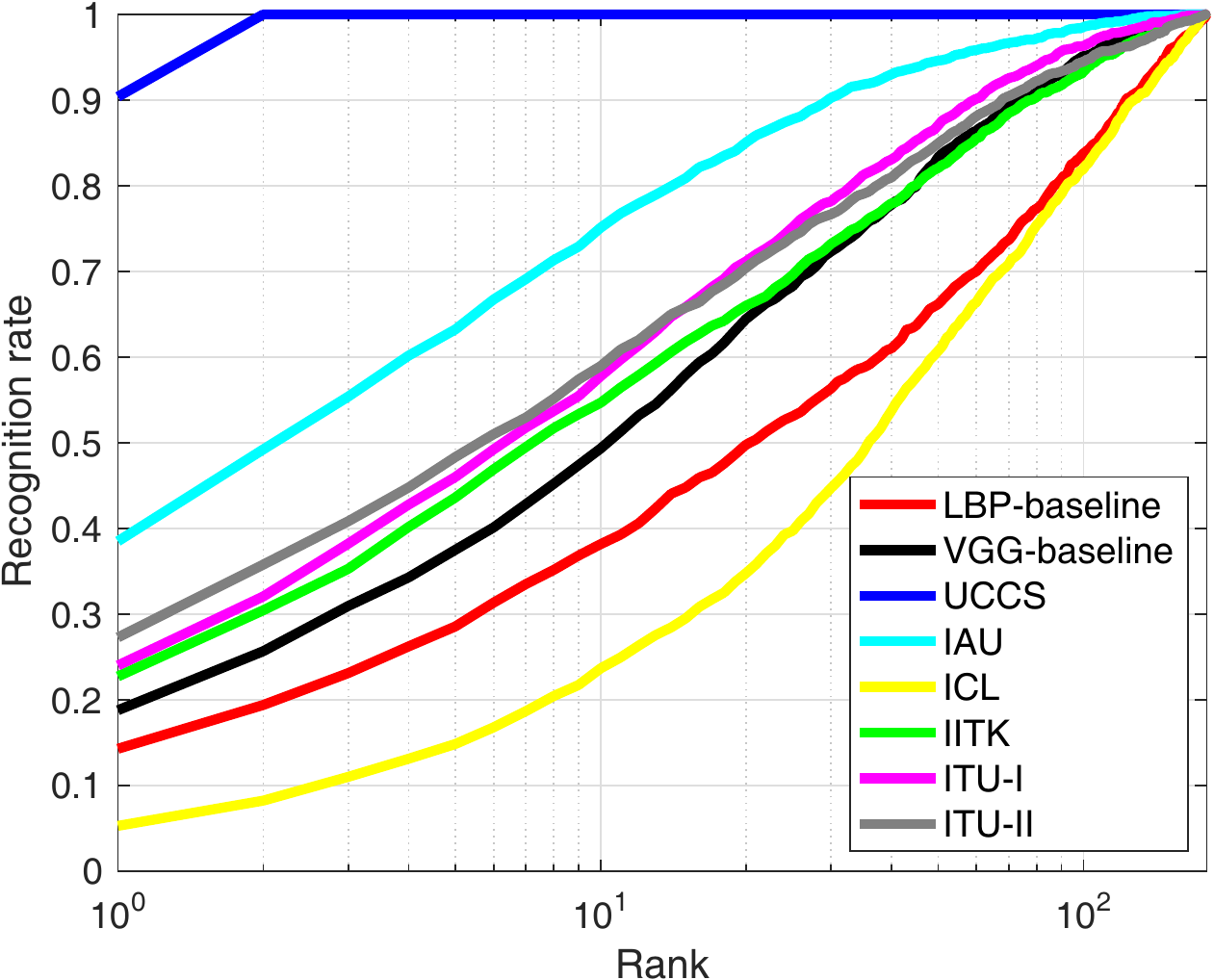}\vspace{1mm}
\caption{CMC curves of the comparative assessment on the images from the AWE dataset. The rank of the experiments is plotted on logarithmic scale to better highlight the performance at the lower ranks. The figure is best viewed in color.}
\label{Fig: AWE comparison}
\end{figure}
\begin{table}[!htb]
\renewcommand{\arraystretch}{1.15}
\centering
\setlength\tabcolsep{9.5pt}
\caption{Comparative evaluation on probe images originating from the AWE dataset. The table shows Rank-1, Rank-5 and AUC values for all tested techniques.}
\vspace{1mm}
\label{Tab: mainResults}
\footnotesize
\begin{tabular}{l rrrr}
\hline
Approach	&	Rank-$1$ (in \%)	&	Rank-$5$ (in \%)	&	AUC \\
\hline
UCCS	&	$90.4$	&	$100$		&	$0.99$4	\\
ICL		&	$5.3$		&	$14.8$	&	$0.717$	\\
IAU		&	$38.5$	&	$63.2$	&	$0.940$	\\
IITK 	& 	$22.7$	& 	$43.6$ 	& 	$0.861$\\
ITU-I	&	$24.0$	&	$46.0$	&	$0.890$	\\
ITU-II	&	$27.3$	&	$48.3$	&	$0.877$	\\
\hline
LBP-baseline	&	$14.3$	&	$28.6$	&	$0.759$	\\
VGG-baseline	&	$18.8$	&	$37.5$	&	$0.861$	\\
\hline

\end{tabular}
\vspace{-3mm}
\end{table}

\textbf{ITU-I}: The ITU-I approach used a $16$-layer VGG model pretrained on the ImageNet dataset and fine-tuned on the UERC training images. Since CNN-based models, such as VGG, need significantly more training data than is available with the UERC dataset to ensure competitive recognition performance, data augmentation was performed and a total of $250,000$ ears were generated from the initial set of $2,304$ ear images. For data augmentation rotation, translation, flipping, intensity changes, cropping, scaling, and sharpening were used. Once the model was fine-tuned, the output of the first fully connected layer ($FC6$) of the VGG model was selected as the ear descriptor.

During run-time, each probe image and its flipped version were processed by the fine-tuned VGG model and the corresponding ear descriptors were matched against the given gallery descriptor with the chi-square distance. Then z-score normalization was performed. This procedure resulted in two scores that were  summed up to produce the final similarity for the probe-to-gallery comparison.

\textbf{ITU-II}: The second approach of the ITU group exploited learned as well as hand-crafted ear descriptors and a fusion procedure applied at the matching score level. Specifically, the ITU-II approach relied on the VGG-based model described above (i.e., see ITU-I) and the LBP-baseline provided by the UERC organizers. Descriptors computed with the two techniques were matched separately against the corresponding gallery descriptors using the chi-square distance, followed by z-score normalization, and resulted in two match scores. The matching procedure was then repeated with flipped probe images and produced another pair of scores. The final probe-to-gallery similarity was ultimately generated by summing up the four scores produced during matching.

\begin{figure}[!tb]
  \centering
  \includegraphics[width=0.9\columnwidth]{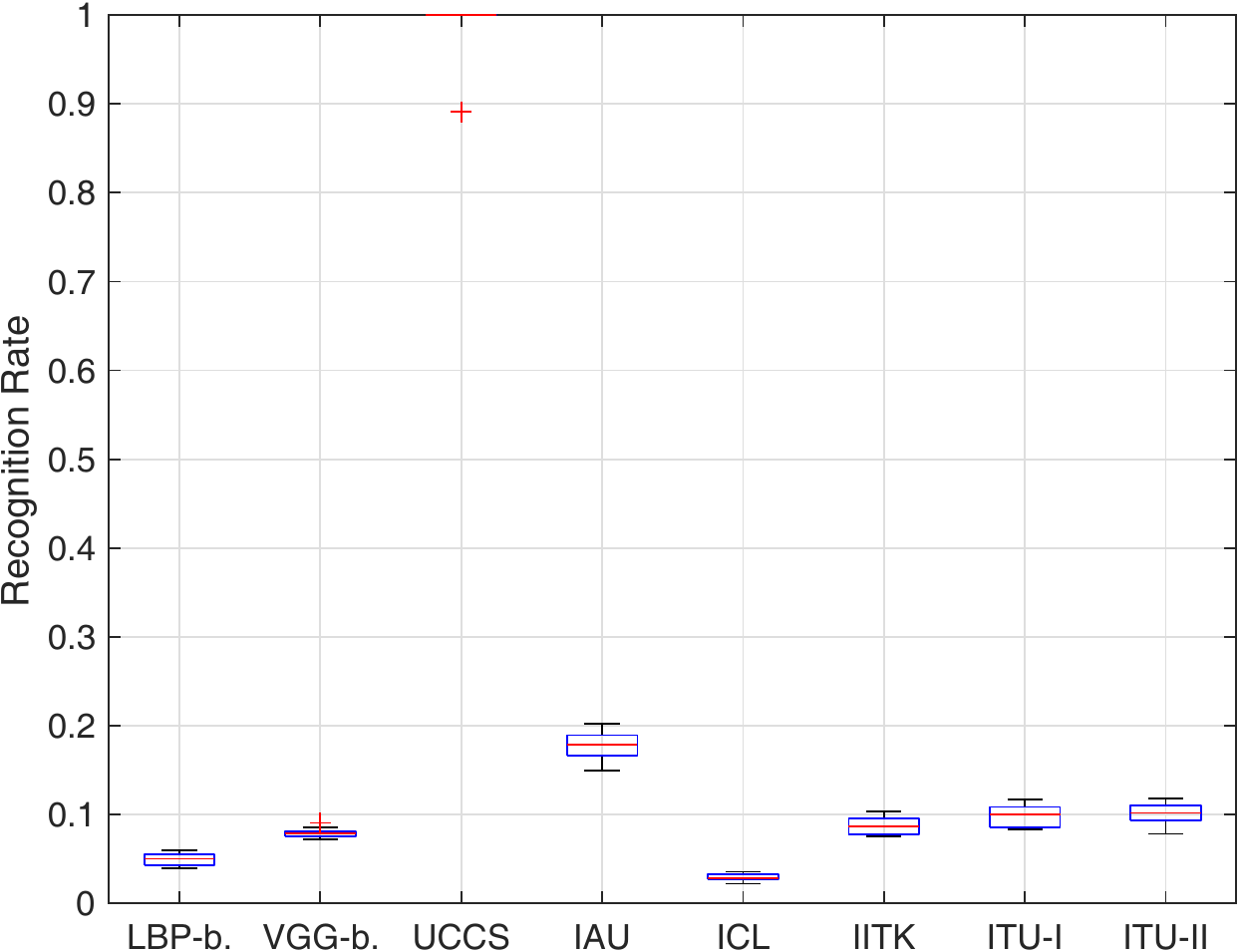}\vspace{2mm}
\caption{Recognition performance with a single gallery image per subject. The box plots show the distribution of the rank-1 recognition rates computed over $10$ experimental runs.}
\label{Fig.: Box-plots}
\end{figure}

\subsection{Summary of participating approaches}

A high-level overview of all participating approaches is given in Table~\ref{Tab: method comparison}. The majority of participating groups approached the challenge with deep learning techniques despite the availability of a relatively small amount of training data. Moreover, among the five deep-learning approaches, four relied on the VGG model architecture but used different preprocessing (e.g., alignment, flipping) and descriptor extraction strategies (e.g., output layer selection). A single approach (ICL) used another architecture, i.e., Inception-ResNet. Hand-crafted descriptors were less popular. Only the UCCS group submitted an approach based on chainlets, while a second one was provided by the challenge organizers in the form of the LBP-baseline. 


\section{Experiments and results}\label{Sec: exper}

In this section we present a comprehensive analysis of all participating approaches and the results of the challenge. 
For potential updates and additions to the results, the reader is referred to the UERC website\footnote{\url{http://awe.fri.uni-lj.si/uerc}}.  
\begin{figure*}[!tb]
  \centering
  \includegraphics[width=1\textwidth]{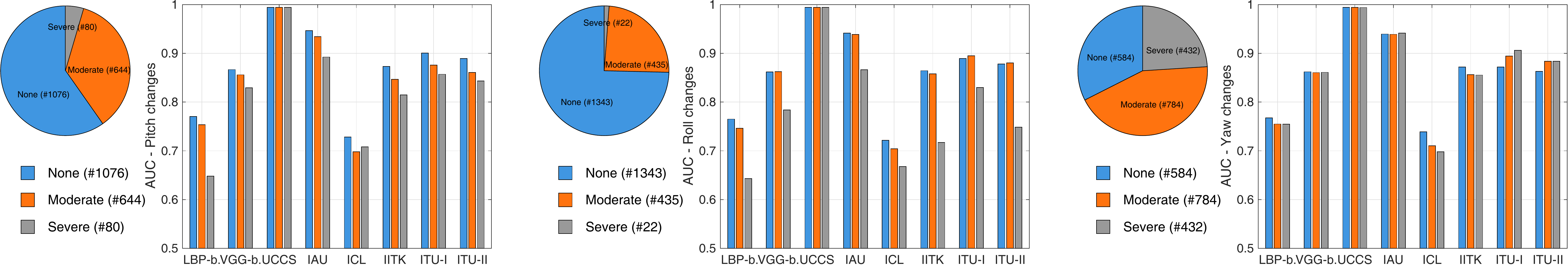}\vspace{2mm}
\caption{Effect of head rotation on ear recognition performance. The left graph shows the change in the AUC with respect to increasing pitch angles, the middle graph shows the AUC change with respect to increasing roll angles and the right graph shows the change in the AUC with respect to increasing yaw angles. The pie charts show the distribution of the images across the different rotation-angle labels. Pitch and roll angles have an adverse effect on the recognition performance of most techniques, whereas yaw angles affect the performance to a lesser extent.}\vspace{-3mm}
\label{Fig: angles}
\end{figure*}

\textbf{Overall performance comparison:} We first investigate how the submitted algorithms perform on ear images captured \textit{in the wild} and how they compare among each other. For this part of the evaluation, we use only $1,800$ probe images originating from the AWE dataset and compute our results on a similarity matrix of size $1,800\times  1,800$ in an all-vs-all experimental setup. Each of the $180$ subjects involved in this experiments is represented in the gallery with $10$ ear images and retrieving any of these $10$ images based on the given probe is counted as correct recognition attempt. 

The results of this experiment are presented in the form of CMC curves in Fig.~\ref{Fig: AWE comparison} and with different performance metrics in Table~\ref{Tab: mainResults}. Overall the UCCS approach results in the best performance with a  recognition rate of $90.4\%$  at a tank of one ($100\%$ at a rank of $2$), followed in order by the IAU, ITU-1, ITU-II, IITK and ICL approaches. A similar ranking can be established if the AUC values are considered instead of the rank-$1$ recognition rates. The LBP and VGG baselines achieve a rank-$1$ recognition rate of $14.3\%$ and $18.8\%$, respectively, with the deep learning baseline having a slight advantage over the hand-crafted LBP descriptor. 

\textbf{Number of galleries per subject:} We next study the effect of reducing the number of gallery images for each subject from $10$ to $1$. This represents a significantly harder problem than in the first experiment, as only a single comparison is available per subject to make an identification decision. Because the $10$ gallery images that are available in the AWE dataset for each subject are divided among left and right ears, comparisons in this experiment may include comparisons of ears from the opposite sides of the head. We perform the experiment $10$-times, so that each of the  gallery images that available in the dataset per subject is used once. The probe set consists of all $1,800$ AWE images. 

As we can see from the box plots in Fig.~\ref{Fig.: Box-plots}, the performance for most of the approaches is halved (on average), except for the UCCS approach, which achieves a rank-$1$ recognition rate of $1$ ($100\%$) in $9$ out of the $10$ experimental runs. These results suggest that having multiple images (of left and right ears) per subject  is detrimental for the recognition performance of most techniques. Even if techniques detect whether the probe and gallery images are from the same side of the head, it is not necessary possible to match the right ear to the left and vice versa. As pointed out by previous studies, e.g.,~\cite{abaza2010towards,yan2005empirical} ears are not always bilateral symmetric, though this is true for most subjects.

\textbf{Head rotation:} Images from the AWE dataset contain annotations with respect to pitch, roll and yaw angles that can be used to explore the impact of head rotation on ear recognition performance. The available annotations (see~\cite{ZigaSurvey2017} for details) are grouped into three categories, i.e., \textit{None}, \textit{Moderate} and \textit{Severe} according to the extent of the head-rotation. Recognition experiments are then performed with probe images of a single category at the time, while the gallery is kept unchanged (i.e., all $1,800$ gallery images are used). As we can see from Fig.~\ref{Fig: angles}, where the AUC values obtained during the experiments are shown, both pitch and roll angles have an adverse affect of most of the techniques, while differences in yaw angles affect the performance to a lesser extent. This results indicates that resampling the ear images to a common size already compensates for the difference in yaw angles, while roll and pitch angles would need to be compensated for explicitly. The UCCS approach is the most robust and is not affected by head rotation.        

\begin{figure}[!tb]
\begin{minipage}{0.49\columnwidth}
  \centering
  \includegraphics[width=1\textwidth]{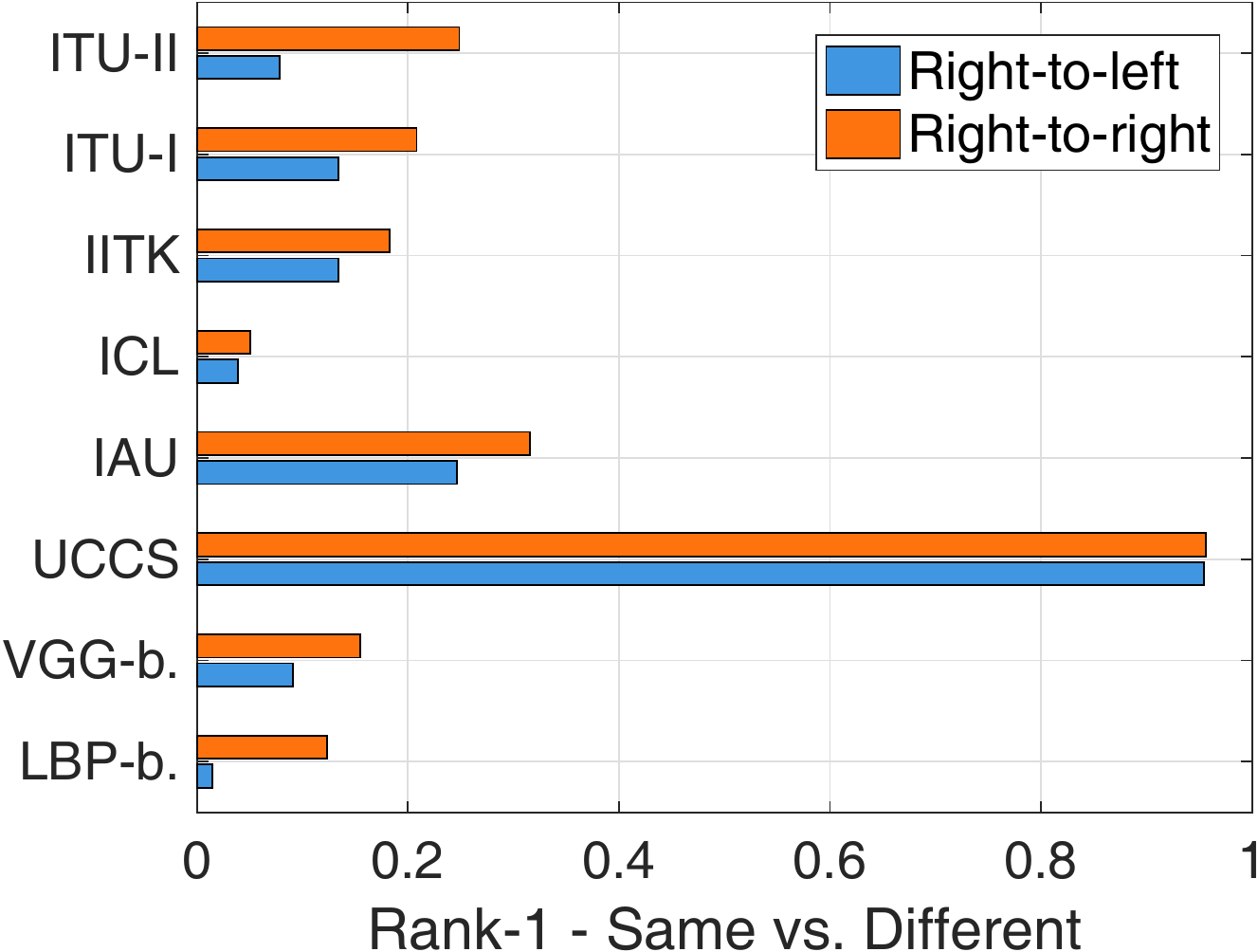}
\end{minipage}
\hfill
\begin{minipage}{0.49\columnwidth}
  \centering
  \includegraphics[width=1\textwidth]{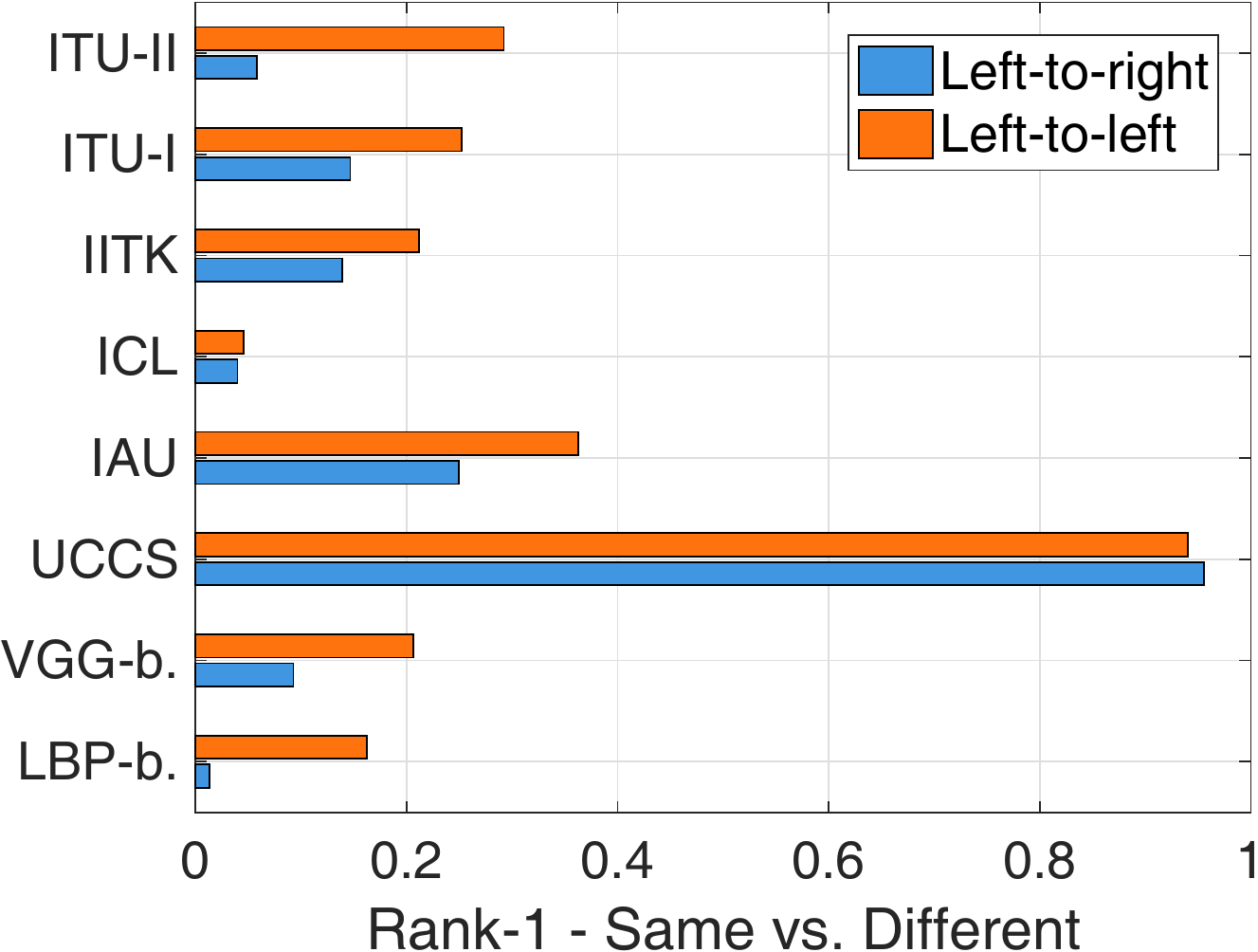}
\end{minipage}\vspace{2.5mm}
\caption{Effect of same-side vs. opposite-side matching. The right graph shows the rank-1 recognition rates when matching right ear probes to either right or left ear galleries. The left graph shows results for the same experiment, but for left-ear probes.}\vspace{-3mm}
\label{Fig: same_diff}
\end{figure}

\textbf{Same-side vs. opposite-side matching:} As already indicated above, the images of all subject in the AWE dataset are split between right and left ears (i.e., images are not side-normalized). When matching probes to galleries, some of the participating approaches try to  detect explicitly whether they are processing left or right ears and then flip the images to a common reference. To evaluate how this process effects performance, we conduct two types of experiments: \textit{i)} experiments with ear images from the same side (e.g., left-to-left), and \textit{ii)} experiments with ear images from opposite sides of the head (e.g., right-to-left). 

The results in Fig.~\ref{Fig: same_diff} show that the SDM fitting approach from the ICL group is the most successful at determining which side of the head the ear images came from, as the performance change in the two experiments is marginal for the ICL approach. The strategies from ITU and IITK are less successful and still result in observable performance drops, but the difference in performance is less than with the baseline techniques, where no effort is made to distinguish left from right ears. The UCCS approach also shows a high level of robustness as a consequence of the exploited chainlet-based descriptor. 
\begin{figure}[!tb]
\begin{minipage}{0.49\columnwidth}
  \centering
  \includegraphics[width=0.98\textwidth]{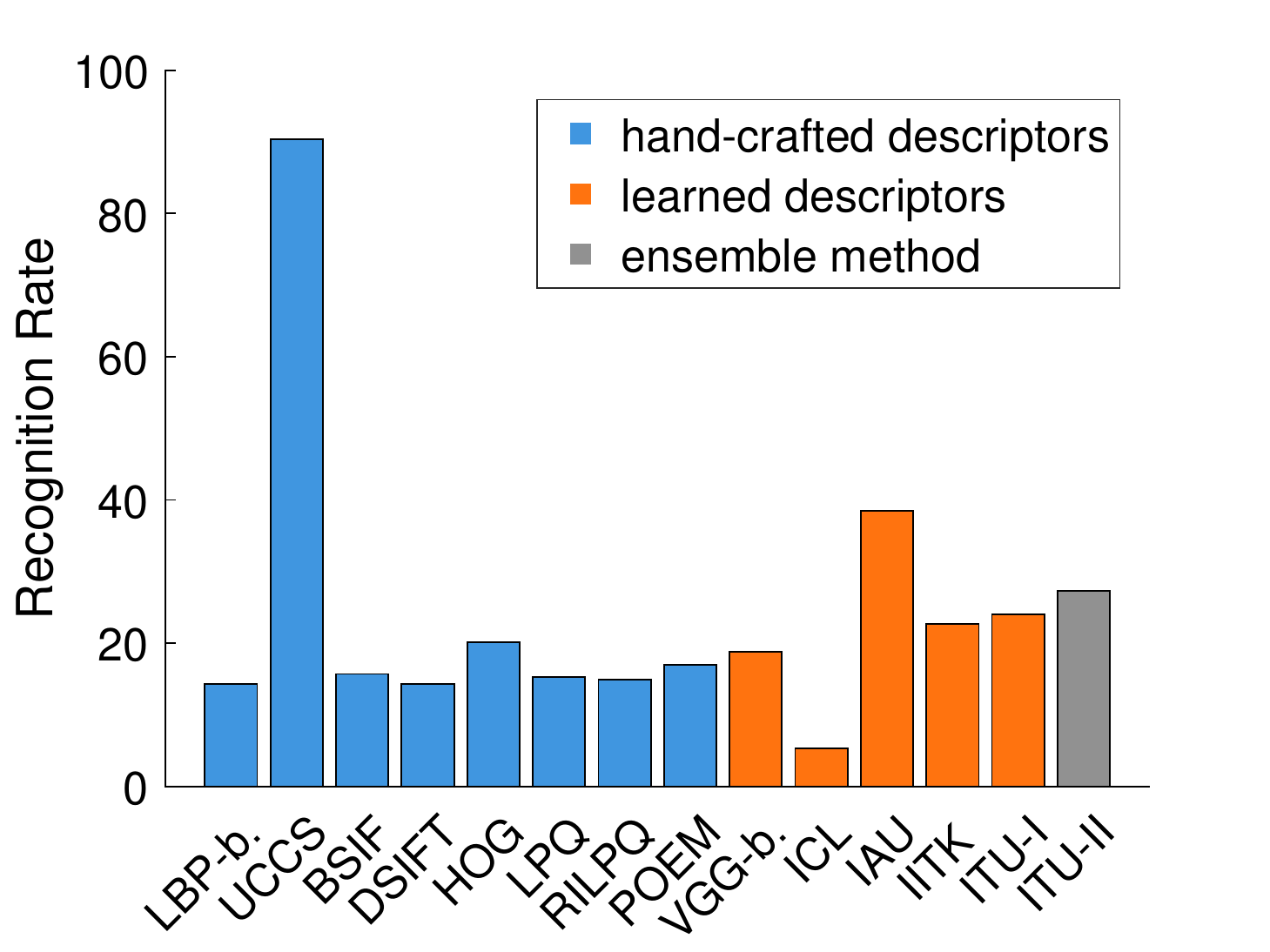}
\end{minipage}
\hfill
\begin{minipage}{0.49\columnwidth}
  \centering
  \includegraphics[width=0.98\textwidth]{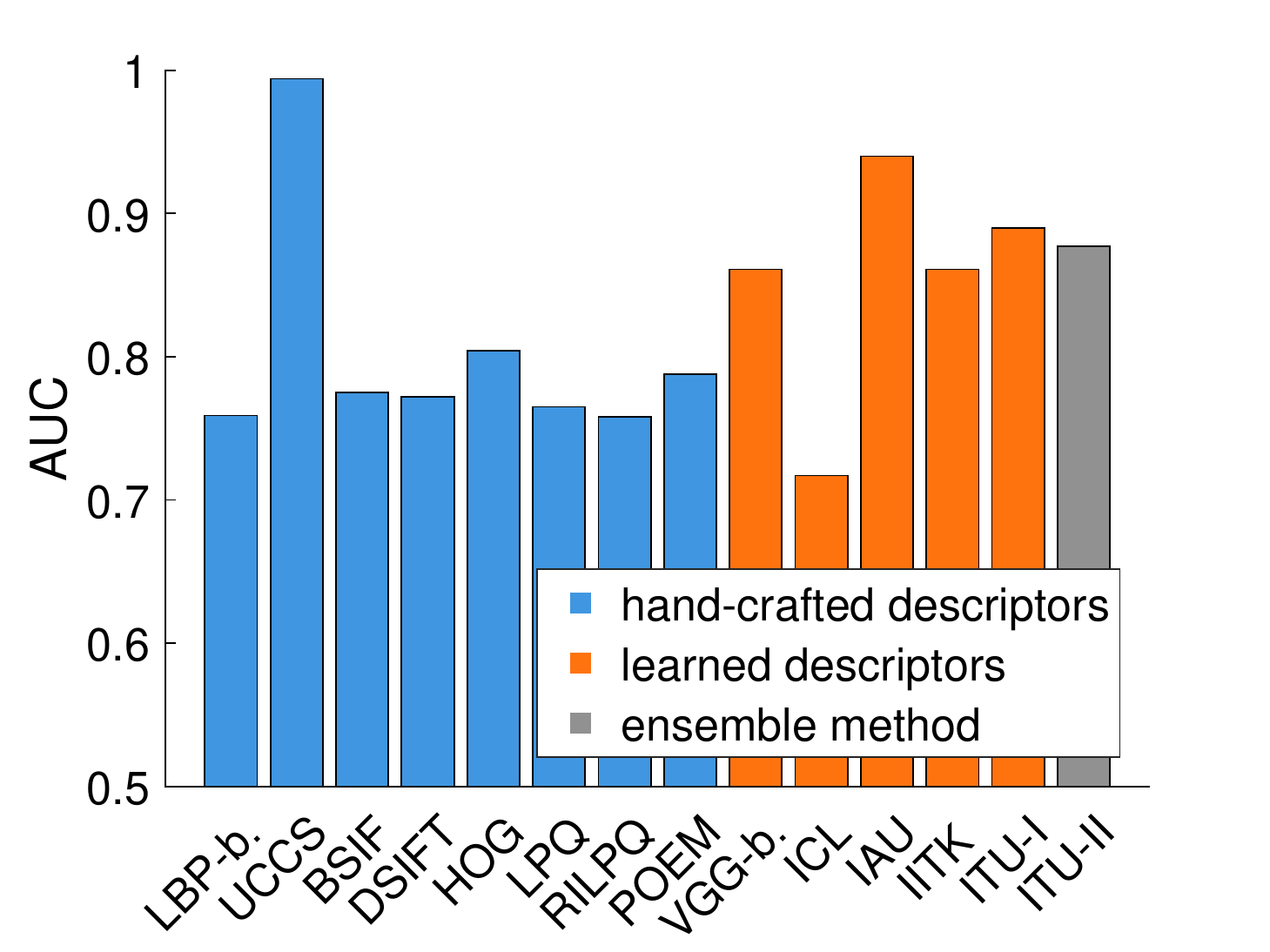}
\end{minipage}\vspace{2mm}
\caption{Comparison of learned and hand-crafted descriptors. The left graph shows the rank-$1$ recognition rate and the right graph shows the AUC of the experiments with $1800$ probe and gallery images of the AWE dataset.}
\label{fig: r1vsr5}\vspace{-4mm}
\end{figure}

\textbf{Learned vs. hand-crafted descriptors:} In our next experiment we compare deep-learning approaches that try to learn ear descriptors from data to techniques that exploit so-called hand-crafted descriptors. Since only UCCS submitted a technique based on hand-crafted descriptors, we also report results for the remaining UERC baselines that are based on HOG~\cite{dalal2005histograms,Pflug2014a}, LBP~\cite{pietikainen2011local}, BSIF~\cite{Kannala2012}, DSIFT~\cite{Lowe2004}, LPQ~\cite{Ojansivu2008}, RILPQ~\cite{OjansivuRILPQ2008} and POEM~\cite{Vu2010} descriptors. For a detailed description of these techniques please see~\cite{Emersic2017}.

From the results in Fig.~\ref{fig: r1vsr5} we see that most of the techniques build around hand-crafted descriptors result in similar performance with a rank-$1$ recognition rate between $14.3\%$ and $20.1\%$, except for the UCCS approach which achieves a recognition rate of $90.4\%$ at rank one. The deep learning approaches, on the other hand, vary significantly in performance despite the fact that $5$ of them use the deep model. We observe rank-$1$ rates between $5.3\%$ and $38.5\%$. This suggests that the training strategy is of paramount importance when the available training data is limited and has a larger impact on the recognition performance than the model architecture. The only ensemble method (ITU-II) benefits from two sources of information and improves upon the performance of the individual techniques (VGG and LBP), from which it was built.

\begin{figure}
  \centering
  \includegraphics[width=0.9\columnwidth]{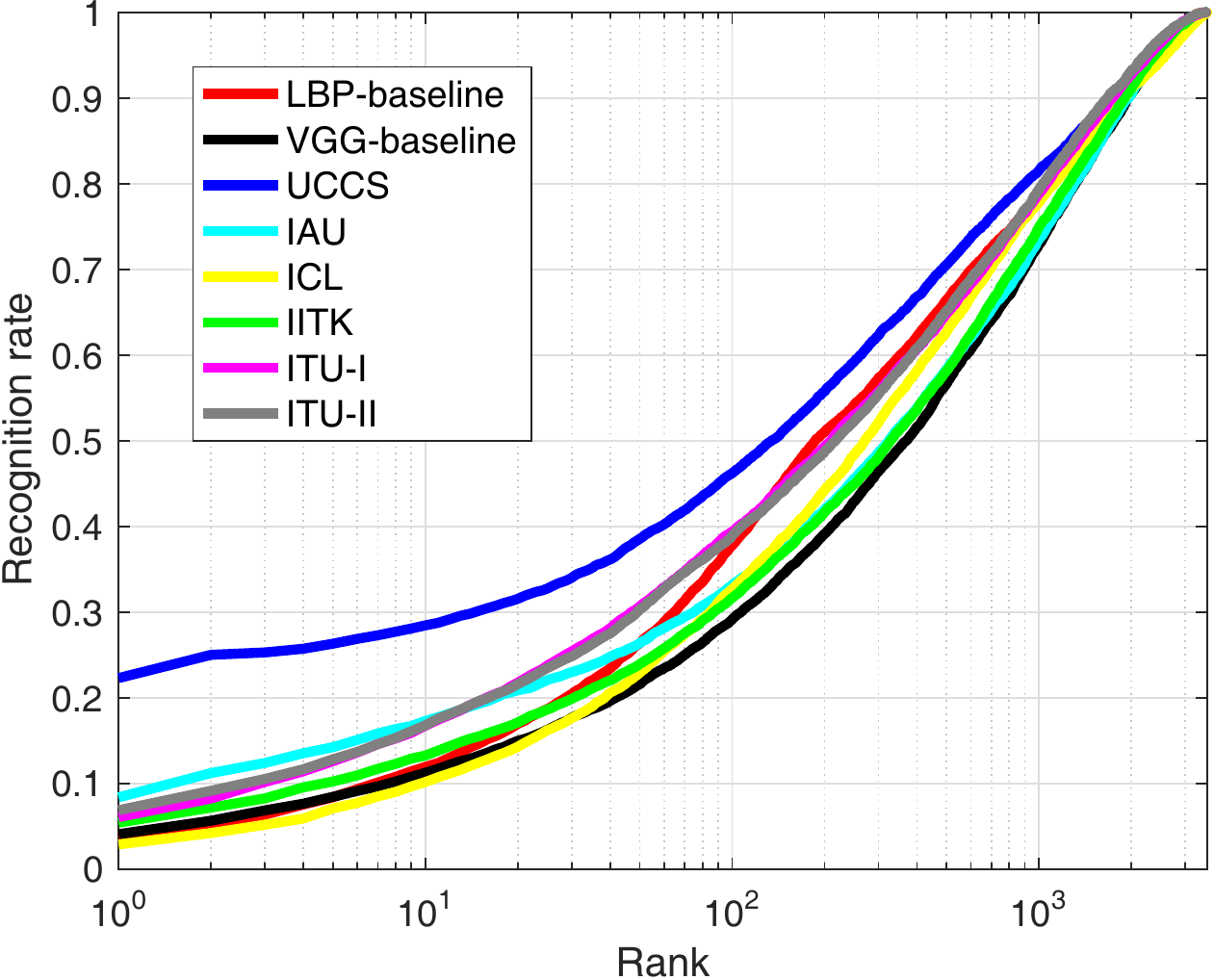}\vspace{1mm}
\caption{CMC curves of the scalability experiments on the entire UERC dataset. The rank of the experiments is plotted on a logarithmic scale to better highlight the performance at the lower ranks. The figure is best viewed in color.}
\label{Fig: scalability}
\end{figure}

\textbf{Scalability:} In our last experiment we evaluate how the recognition techniques 
scale with larger probe and gallery sets. We show CMC curves generated based on $7,442$ probe images belonging to $1,482$ subjects and $9,500$ gallery images of $3,540$ subjects in Fig.~\ref{Fig: scalability}. Note that the gallery also contains identities that are not in the probe set and act as distractors (to use the terminology from~\cite{kemelmacher2016megaface}) for the recognition techniques. Quantitative performance metrics of the experiments are presented in Table~\ref{Tab: mainResultsScale}. All techniques deteriorate significantly in performance with scale. The best performer is again the UCCS approach with a recognition rate of $22.3\%$ at a rank of $1$, followed in order by the IAU, ITU-II, ITU-I, IITK and ICL approaches. When the AUC values are considered the ITU approaches become competitive and get close to the performance of the UCCS technique due to a better performance at the higher ranks. 

\begin{table}[!bt]
\renewcommand{\arraystretch}{1.12}
\centering
\setlength\tabcolsep{9.5pt}
\caption{Key performance metrics of the scalability experiments. The table shows Rank-1, Rank-5 and AUC values for all tested techniques.}
\vspace{1mm}
\label{Tab: mainResultsScale}
\footnotesize
\begin{tabular}{l rrrr}
\hline
Approach	&	Rank-$1$ (in \%)	&	Rank-$5$ (in \%)	&	AUC \\
\hline
UCCS	&	$22.3$		&	$26.3$		&	$0.858$	\\
ICL		&	$2.9$		&	$7.08$	&	$0.823$	\\
IAU		&	$8.4$		&	$14.2$	&	$0.810$	\\
IITK 	& 	$5.4 $		& 	$10.2$ 	& 	$0.814$\\
ITU-I	&	$6.1$		&	$12.6$	&	$0.842$	\\
ITU-II	&	$6.9$		&	$12.8$	&	$0.844$	\\
\hline
LBP-baseline	&	$3.75$	&	$8.38$	&	$0.835$	\\
VGG-baseline	&	$4.07$	&	$8.43$	&	$0.804$	\\
\hline
\vspace{-5mm}
\end{tabular}
\end{table}


\section{Conclusion}\label{Sec: conclusion}

We have presented the results of the first Unconstrained Ear Recognition Challenge (UERC) that aimed at evaluating the current state of technology in the field of ear recognition from images captured in unconstrained environments. While additional experiments are presented in the Appendix, open questions still remain for future challenges. 

Several important findings were made in this paper, e.g.: \textit{i)} significant performance improvements are needed before the ear-recognition technology is suitable for deployment in unconstrained environments at scale, \textit{ii)}  existing approaches are mostly sensitive to specific head rotations (in pitch and roll directions, but not yaw), and \textit{iii)} it is detrimental to have multiple images of both ears in the gallery, as identity inference based on a single-image-per-subject basis leads to poor recognition performance.  



{\small
\bibliographystyle{ieee}
\bibliography{refs}
}

\newpage
\appendix
\label{Sec: appendix}
\section{Additional UERC results}

In this appendix we present additional results from the UERC. The results relate to the impact of occlusions, gender and image size on performance. Qualitative results are also presented.  

\begin{figure}[!b]
\begin{minipage}{0.49\columnwidth}
  \centering
  \includegraphics[width=1\textwidth]{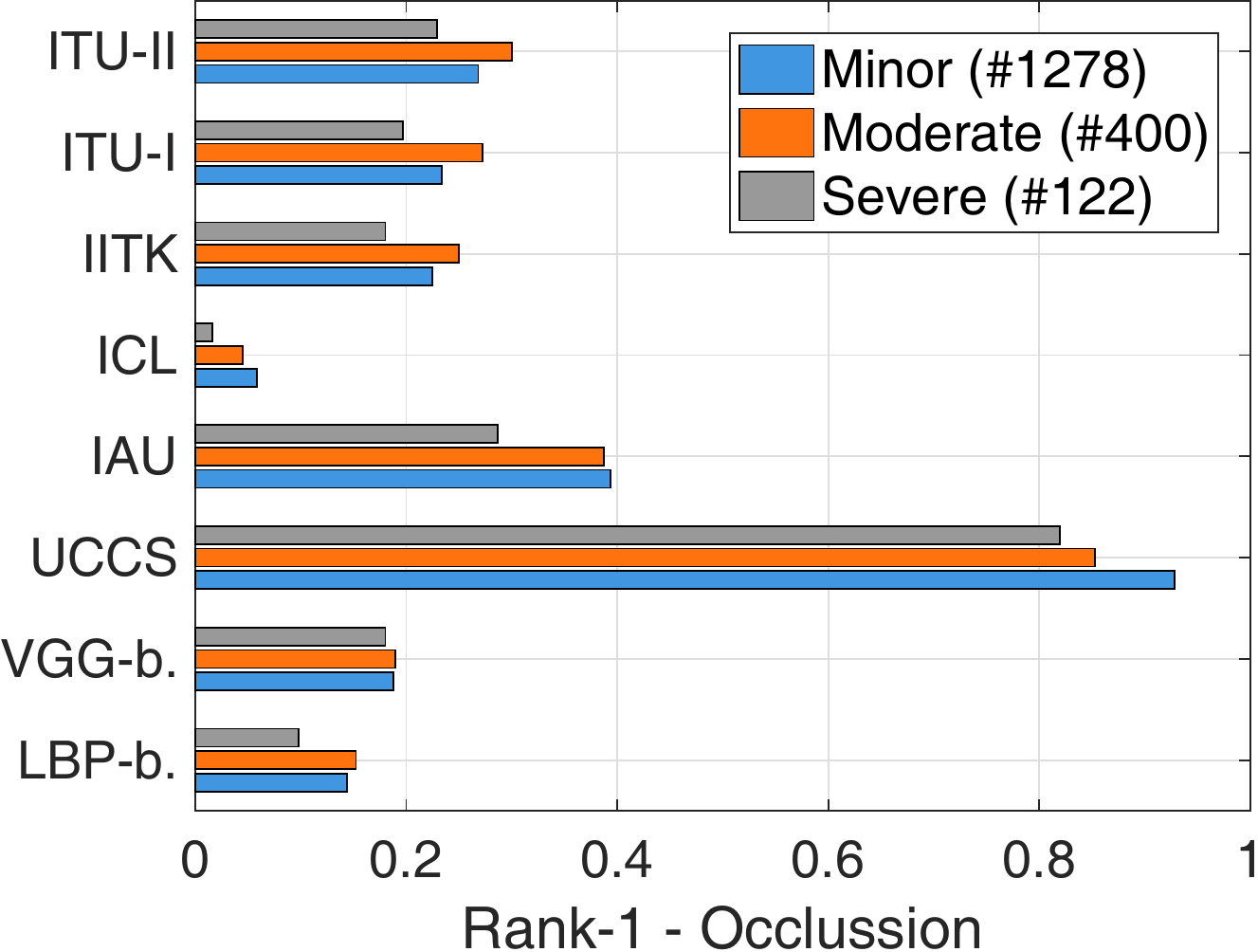}
\end{minipage}
\hfill
\begin{minipage}{0.49\columnwidth}
  \centering
  \includegraphics[width=1\textwidth]{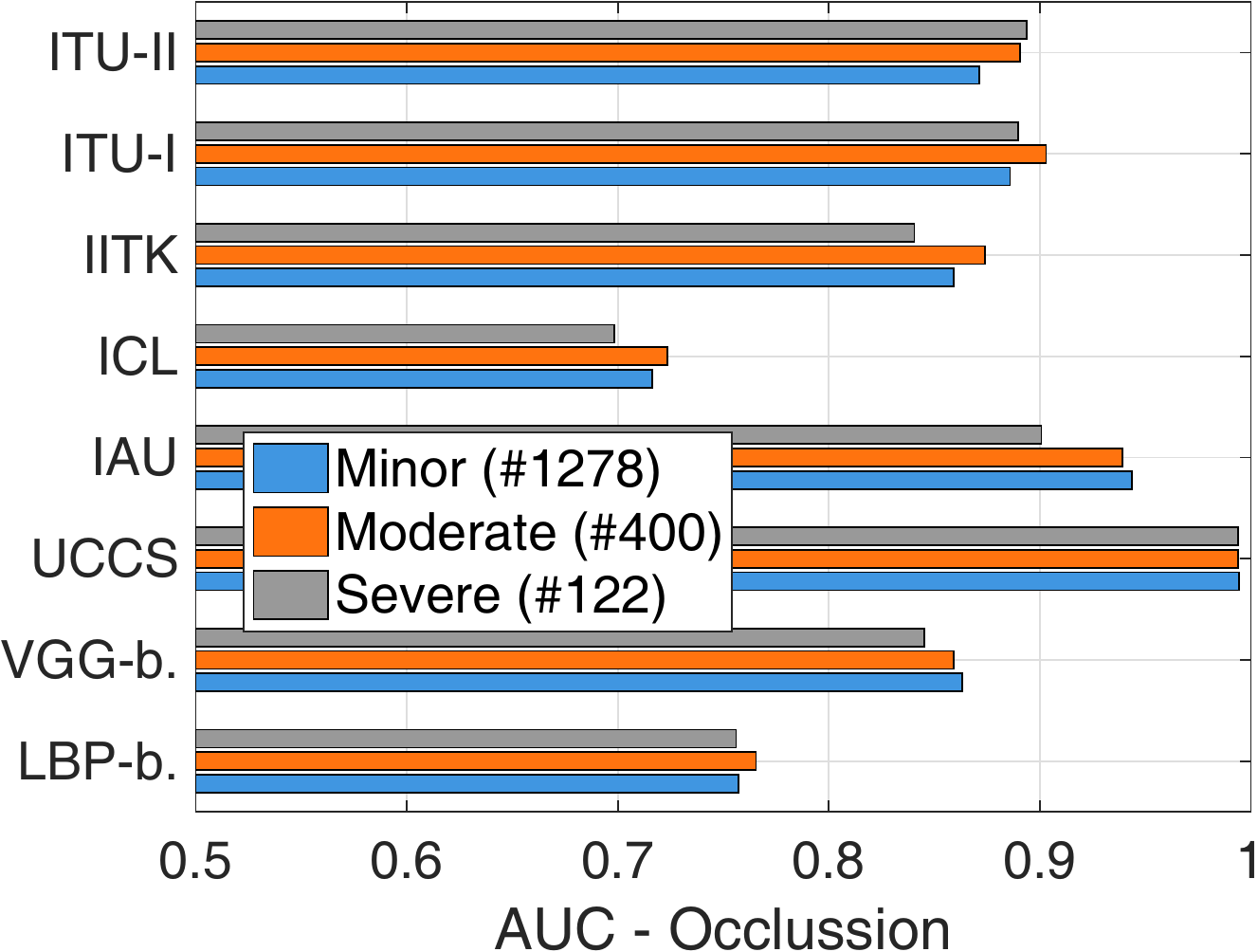}
\end{minipage}\vspace{2.5mm}
\caption{Impact of occlusions on the recognition performance. The left graph shows the rank-$1$ recognition rates and the right graph shows AUC values for all tested techniques. The number in the brackets indicate the number of images for each label.}\vspace{-3mm}
\label{Fig: occlu}
\end{figure}

\textbf{Occlusion:} We explore the impact of partial ear occlusion on images originating from the AWE dataset. We run three experiments and during each use only probe images labeled with one of the following occlusion labels: \textit{Minor}, \textit{Moderate} and \textit{Severe}. For the gallery we use all $1,800$ AWE images from the UERC test partition. The results for this experiment are presented for the rank-$1$ recognition rate in Fig.~\ref{Fig: occlu} (left) and for the AUC in Fig.~\ref{Fig: occlu} (right). As can be seen, minor and moderate occlusions caused, for example, by hair have limited effect on the recognition performance, whereas severe occlusion from scarfs, large accessories and a like have a larger impact on all submitted approaches.  

\begin{figure}[!h]
  \centering
  \includegraphics[width=0.9\columnwidth]{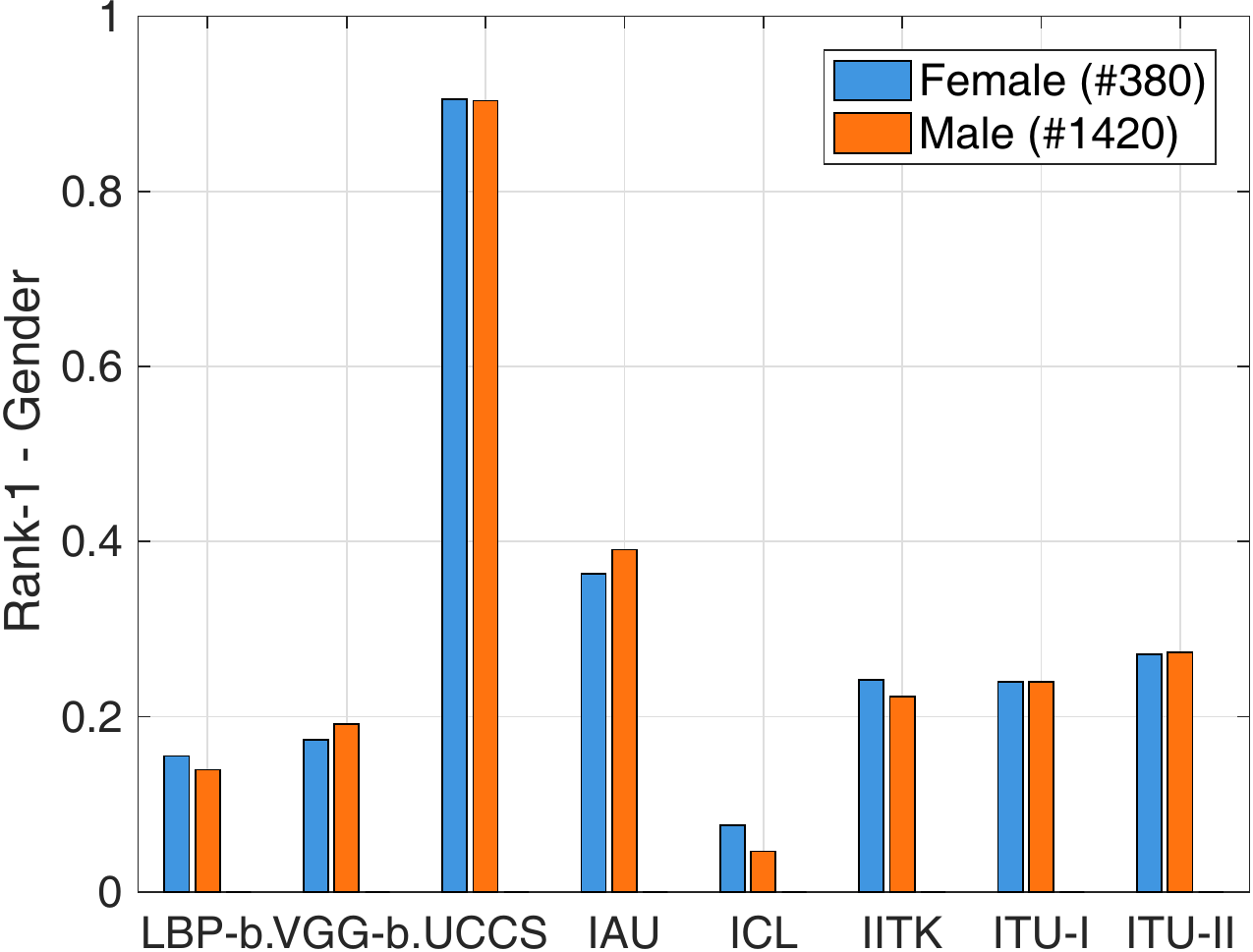}\vspace{1mm}
\caption{Impact of gender on the recognition performance. The  graph shows the rank-$1$ recognition rates for all tested techniques. The number in the brackets indicate the number of images for each label.}
\label{Fig: gender}
\end{figure}

\textbf{Gender:} To study the effect of gender on the recognition performance, we run separate test with probe images from the AWE dataset belonging to male ($1,420$ images)  and female  subjects ($320$ images). The gallery again comprises all $1,800$ available images. We can see from the rank-$1$ recognition rates in Fig.~\ref{Fig: gender} that the performance of all approaches participating in the UERC is more or less unaffected by gender. Both men and women are recognized equally well.

\begin{figure}[!t]
\begin{minipage}{0.49\columnwidth}
  \centering
  \includegraphics[width=1\textwidth]{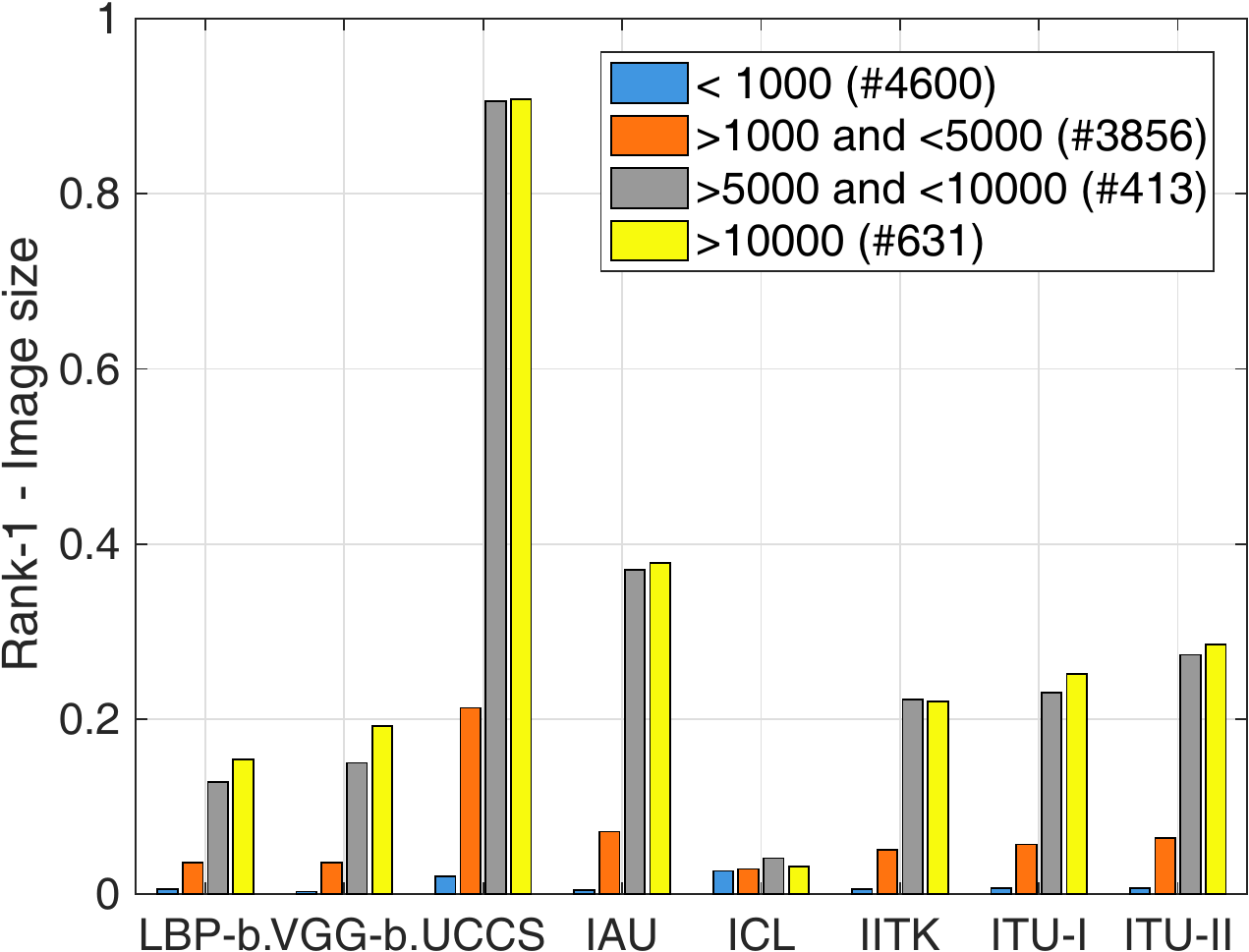}
\end{minipage}
\hfill
\begin{minipage}{0.49\columnwidth}
  \centering
  \includegraphics[width=1\textwidth]{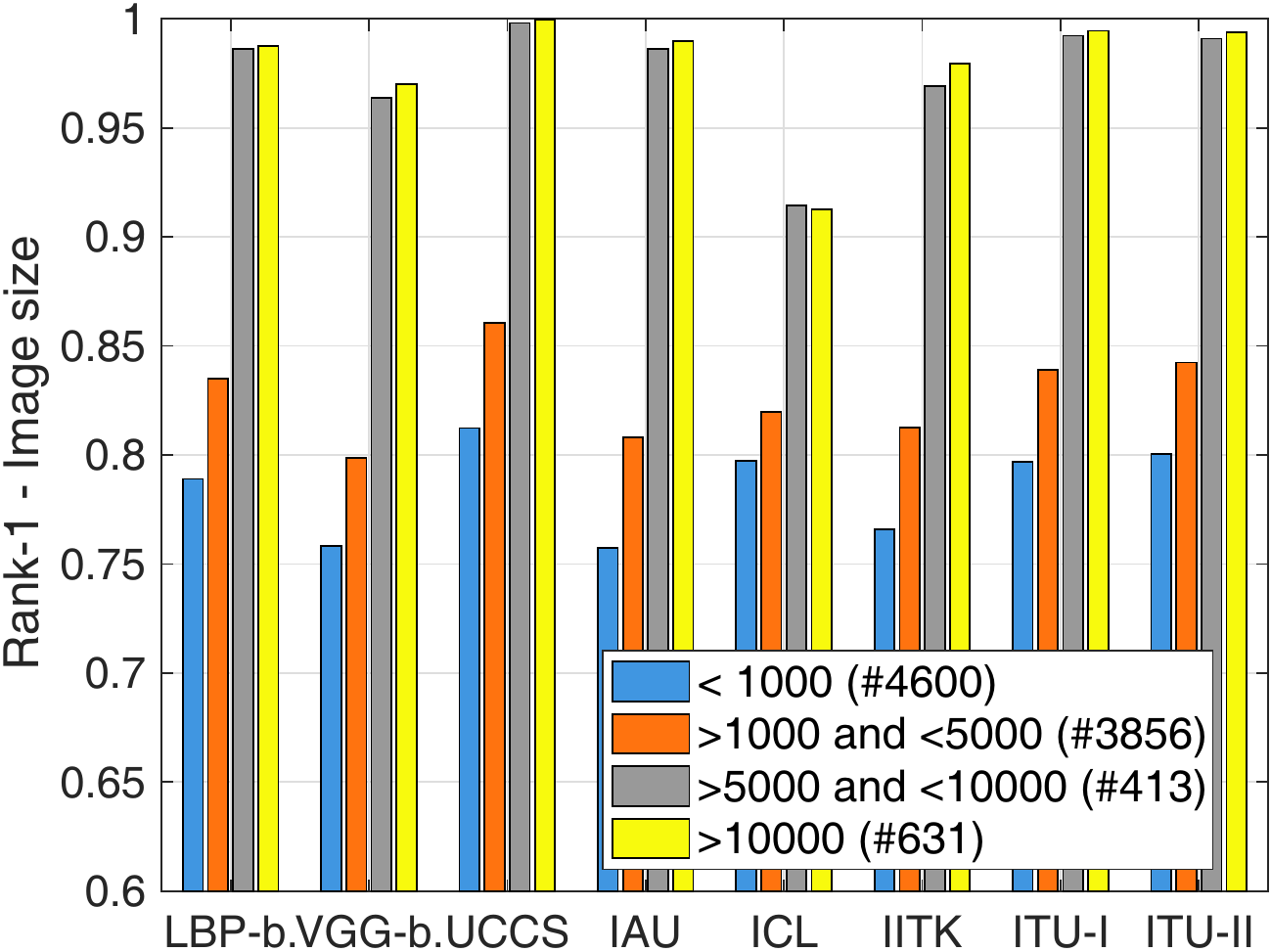}
\end{minipage}\vspace{2.5mm}
\caption{Impact of image size on the recognition performance. The left graph shows the rank-$1$ recognition rates and the right graph shows AUC values for all tested techniques. The number in the brackets indicate the number of images for each label.}\vspace{-3mm}
\label{Fig: resolution}
\end{figure}

\textbf{Image size:} Next, we explore the impact of image size on the recognition performance. For this experiment we use the entire similarity matrix and select probe images from four categories: \textit{i)} images with a pixel count below $1,000$ ($4,600$ images), \textit{ii)} images of size between $1,000$ and $5,000$ pixels ($3,856$ images), \textit{iii)} images of size between $5,000$ and $10,000$ pixels ($413$ images), and \textit{iv)} images with more than $10,000$ pixels ($631$ images). We use all $9,500$ gallery images for the assessment and present the results in the form of rank-$1$ recognition rates and AUC values in Fig.~\ref{Fig: resolution}.

As we can see, all approaches are affected by the changes in image size. The biggest performance drop is seen for the UCCS approach, which performs very well on the large images from the AWE dataset, but is less successful on the smaller images of the newly collected part of the UERC data. Also interesting are the results for the ICL  approach, which show that the rank-$1$ recognition rate is close for all image-size categories. The remaining approaches all behave  similarly with the performance decreasing with decreasing image size. This observation is important and shows that images of sufficient size need to be available for ear recognition. Alternatively, models robust to image size or supper-resolution techniques may need to be incorporated into the recognition pipelines to make ear recognition work with small images.    

\begin{table*}[!htb]
\renewcommand{\arraystretch}{1.1}
\centering
\caption{Qualitative examples of the recognition performance. The table shows a few selected probe images and the best and second best match from the gallery. The correct gallery and information about the rank at which the correct gallery was retrieved is also given. The table shows some of the errors and corrected identification attempts made by the tested approaches.}
\vspace{1mm}
\label{Tab: qualitative}
\resizebox{\textwidth}{!}{%
\footnotesize
\begin{tabular}{l|ccccc |ccccc}

Approach 						& Probe	&	$1$. match 	&	$2$. match	&	Correct  & Retrieved at & Probe	&	$1$. match 	&	$2$. match	&	Correct  & Retrieved at\\
\hline 
\multirow{2}{*}{LBP-baseline} 	&  \includegraphics[width=0.07\textwidth,height=0.07\textwidth]{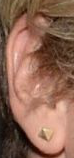} & \includegraphics[width=0.07\textwidth,height=0.07\textwidth]{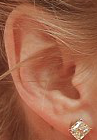} & \includegraphics[width=0.07\textwidth,height=0.07\textwidth]{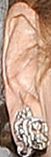}& \includegraphics[width=0.07\textwidth,height=0.07\textwidth]{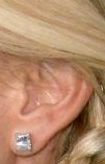}&  rank $39$
								& \includegraphics[width=0.07\textwidth,height=0.07\textwidth]{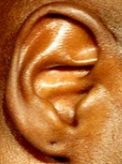} & \includegraphics[width=0.07\textwidth,height=0.07\textwidth]{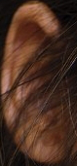} & \includegraphics[width=0.07\textwidth,height=0.07\textwidth]{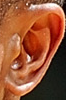}&  \includegraphics[width=0.07\textwidth,height=0.07\textwidth]{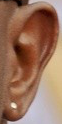}& rank $85$ \\
                    			& \includegraphics[width=0.07\textwidth,height=0.07\textwidth]{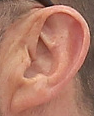}& \includegraphics[width=0.07\textwidth,height=0.07\textwidth]{figures/11_07.png}  & \includegraphics[width=0.07\textwidth,height=0.07\textwidth]{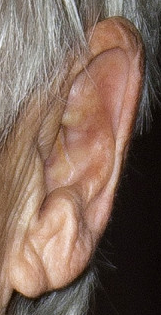}&  \includegraphics[width=0.07\textwidth,height=0.07\textwidth]{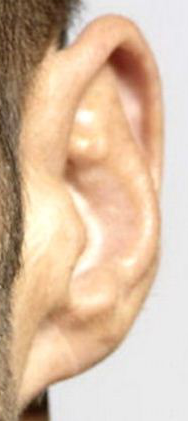}& rank $15$
                                & \includegraphics[width=0.07\textwidth,height=0.07\textwidth]{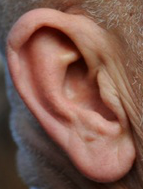}& \includegraphics[width=0.07\textwidth,height=0.07\textwidth]{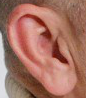} & \includegraphics[width=0.07\textwidth,height=0.07\textwidth]{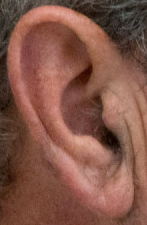}&  \includegraphics[width=0.07\textwidth,height=0.07\textwidth]{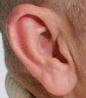}& rank $1$\\ \hline

\multirow{2}{*}{VGG-baseline}   &  \includegraphics[width=0.07\textwidth,height=0.07\textwidth]{figures/s3i10.png} & \includegraphics[width=0.07\textwidth,height=0.07\textwidth]{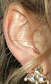}& \includegraphics[width=0.07\textwidth,height=0.07\textwidth]{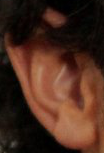}& \includegraphics[width=0.07\textwidth,height=0.07\textwidth]{figures/s3i9.png}&  rank $49$
								& \includegraphics[width=0.07\textwidth,height=0.07\textwidth]{figures/s6i1.png} & \includegraphics[width=0.07\textwidth,height=0.07\textwidth]{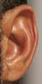} & \includegraphics[width=0.07\textwidth,height=0.07\textwidth]{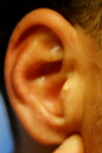}& \includegraphics[width=0.07\textwidth,height=0.07\textwidth]{figures/s6i7.png}& rank $45$\\
                    			& \includegraphics[width=0.07\textwidth,height=0.07\textwidth]{figures/s12i1.png}& \includegraphics[width=0.07\textwidth,height=0.07\textwidth]{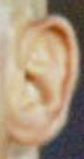} & \includegraphics[width=0.07\textwidth,height=0.07\textwidth]{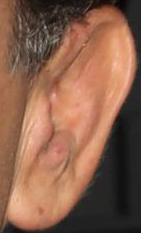}& \includegraphics[width=0.07\textwidth,height=0.07\textwidth]{figures/s12i7.png}& rank $34$ 
                                & \includegraphics[width=0.07\textwidth,height=0.07\textwidth]{figures/s13i2.png}& \includegraphics[width=0.07\textwidth,height=0.07\textwidth]{figures/13_4.png} & \includegraphics[width=0.07\textwidth,height=0.07\textwidth]{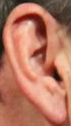}& \includegraphics[width=0.07\textwidth,height=0.07\textwidth]{figures/s13i4.png}& rank $1$\\ \hline
\multirow{2}{*}{UCCS}   		&  \includegraphics[width=0.07\textwidth,height=0.07\textwidth]{figures/s3i10.png} & \includegraphics[width=0.07\textwidth,height=0.07\textwidth]{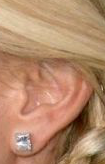}& \includegraphics[width=0.07\textwidth,height=0.07\textwidth]{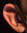}& \includegraphics[width=0.07\textwidth,height=0.07\textwidth]{figures/s3i9.png}&  rank $1$
								& \includegraphics[width=0.07\textwidth,height=0.07\textwidth]{figures/s6i1.png} & \includegraphics[width=0.07\textwidth,height=0.07\textwidth]{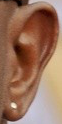} & \includegraphics[width=0.07\textwidth,height=0.07\textwidth]{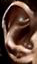}& \includegraphics[width=0.07\textwidth,height=0.07\textwidth]{figures/s6i7.png}& rank $1$\\
                    			& \includegraphics[width=0.07\textwidth,height=0.07\textwidth]{figures/s12i1.png}& \includegraphics[width=0.07\textwidth,height=0.07\textwidth]{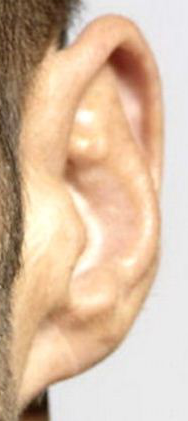} & \includegraphics[width=0.07\textwidth,height=0.07\textwidth]{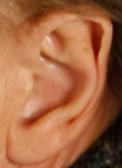}& \includegraphics[width=0.07\textwidth,height=0.07\textwidth]{figures/s12i7.png}& rank $1$ 
                                & \includegraphics[width=0.07\textwidth,height=0.07\textwidth]{figures/s13i2.png}& \includegraphics[width=0.07\textwidth,height=0.07\textwidth]{figures/13_4.png} & \includegraphics[width=0.07\textwidth,height=0.07\textwidth]{figures/8_4.png}& \includegraphics[width=0.07\textwidth,height=0.07\textwidth]{figures/s13i4.png}& rank $1$\\ \hline
\multirow{2}{*}{IAU}   			&  \includegraphics[width=0.07\textwidth,height=0.07\textwidth]{figures/s3i10.png} & \includegraphics[width=0.07\textwidth,height=0.07\textwidth]{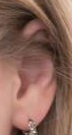}& \includegraphics[width=0.07\textwidth,height=0.07\textwidth]{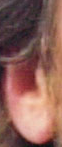}& \includegraphics[width=0.07\textwidth,height=0.07\textwidth]{figures/s3i9.png}&  rank $11$
								& \includegraphics[width=0.07\textwidth,height=0.07\textwidth]{figures/s6i1.png} & \includegraphics[width=0.07\textwidth,height=0.07\textwidth]{figures/18_7.png} & \includegraphics[width=0.07\textwidth,height=0.07\textwidth]{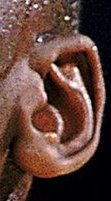}& \includegraphics[width=0.07\textwidth,height=0.07\textwidth]{figures/s6i7.png}& rank $20$\\
                    			& \includegraphics[width=0.07\textwidth,height=0.07\textwidth]{figures/s12i1.png}& \includegraphics[width=0.07\textwidth,height=0.07\textwidth]{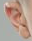} & \includegraphics[width=0.07\textwidth,height=0.07\textwidth]{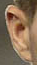}& \includegraphics[width=0.07\textwidth,height=0.07\textwidth]{figures/s12i7.png}& rank $78$
                                & \includegraphics[width=0.07\textwidth,height=0.07\textwidth]{figures/s13i2.png}& \includegraphics[width=0.07\textwidth,height=0.07\textwidth]{figures/13_4.png} & \includegraphics[width=0.07\textwidth,height=0.07\textwidth]{figures/59_4.png}& \includegraphics[width=0.07\textwidth,height=0.07\textwidth]{figures/s13i4.png}& rank $1$\\ \hline
\multirow{2}{*}{ICL}  			&  \includegraphics[width=0.07\textwidth,height=0.07\textwidth]{figures/s3i10.png} & \includegraphics[width=0.07\textwidth,height=0.07\textwidth]{figures/80_9.png} & \includegraphics[width=0.07\textwidth,height=0.07\textwidth]{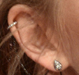}& \includegraphics[width=0.07\textwidth,height=0.07\textwidth]{figures/s3i9.png}&  rank $6$ 
								& \includegraphics[width=0.07\textwidth,height=0.07\textwidth]{figures/s6i1.png} & \includegraphics[width=0.07\textwidth,height=0.07\textwidth]{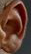} & \includegraphics[width=0.07\textwidth,height=0.07\textwidth]{figures/18_7.png}& \includegraphics[width=0.07\textwidth,height=0.07\textwidth]{figures/s6i7.png}& rank $142$\\
                    			& \includegraphics[width=0.07\textwidth,height=0.07\textwidth]{figures/s12i1.png}& \includegraphics[width=0.07\textwidth,height=0.07\textwidth]{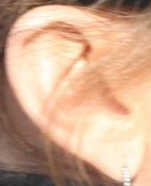} & \includegraphics[width=0.07\textwidth,height=0.07\textwidth]{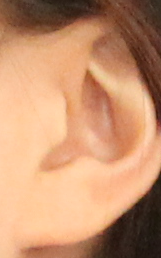}& \includegraphics[width=0.07\textwidth,height=0.07\textwidth]{figures/s12i7.png}& rank $151$
                                & \includegraphics[width=0.07\textwidth,height=0.07\textwidth]{figures/s13i2.png}& \includegraphics[width=0.07\textwidth,height=0.07\textwidth]{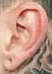}& \includegraphics[width=0.07\textwidth,height=0.07\textwidth]{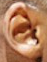}& \includegraphics[width=0.07\textwidth,height=0.07\textwidth]{figures/s13i4.png}& rank $5$\\ \hline
\multirow{2}{*}{IITK}   &  \includegraphics[width=0.07\textwidth,height=0.07\textwidth]{figures/s3i10.png} & \includegraphics[width=0.07\textwidth,height=0.07\textwidth]{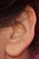}& \includegraphics[width=0.07\textwidth,height=0.07\textwidth]{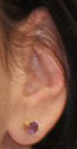}& \includegraphics[width=0.07\textwidth,height=0.07\textwidth]{figures/s3i9.png}& rank $22$ 
								& \includegraphics[width=0.07\textwidth,height=0.07\textwidth]{figures/s6i1.png} & \includegraphics[width=0.07\textwidth,height=0.07\textwidth]{figures/23_7.png}& \includegraphics[width=0.07\textwidth,height=0.07\textwidth]{figures/18_7.png}& \includegraphics[width=0.07\textwidth,height=0.07\textwidth]{figures/s6i7.png}& rank $151$\\
                    			& \includegraphics[width=0.07\textwidth,height=0.07\textwidth]{figures/s12i1.png}& \includegraphics[width=0.07\textwidth,height=0.07\textwidth]{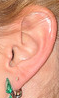} & \includegraphics[width=0.07\textwidth,height=0.07\textwidth]{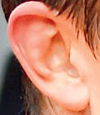}& \includegraphics[width=0.07\textwidth,height=0.07\textwidth]{figures/s12i7.png}& rank $78$
                                & \includegraphics[width=0.07\textwidth,height=0.07\textwidth]{figures/s13i2.png}& \includegraphics[width=0.07\textwidth,height=0.07\textwidth]{figures/13_4.png} & \includegraphics[width=0.07\textwidth,height=0.07\textwidth]{figures/59_4.png}& \includegraphics[width=0.07\textwidth,height=0.07\textwidth]{figures/s13i4.png}& rank $1$\\ \hline
\multirow{2}{*}{ITU-I}  &  \includegraphics[width=0.07\textwidth,height=0.07\textwidth]{figures/s3i10.png} & \includegraphics[width=0.07\textwidth,height=0.07\textwidth]{figures/5_9.png}&\includegraphics[width=0.07\textwidth,height=0.07\textwidth]{figures/149_09.png} &  \includegraphics[width=0.07\textwidth,height=0.07\textwidth]{figures/s3i9.png}&  rank $3$
								& \includegraphics[width=0.07\textwidth,height=0.07\textwidth]{figures/s6i1.png} & \includegraphics[width=0.07\textwidth,height=0.07\textwidth]{figures/18_7.png} & \includegraphics[width=0.07\textwidth,height=0.07\textwidth]{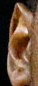}& \includegraphics[width=0.07\textwidth,height=0.07\textwidth]{figures/s6i7.png}& rank $25$\\
                    			& \includegraphics[width=0.07\textwidth,height=0.07\textwidth]{figures/s12i1.png}& \includegraphics[width=0.07\textwidth,height=0.07\textwidth]{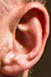} & \includegraphics[width=0.07\textwidth,height=0.07\textwidth]{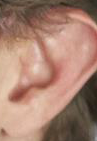}& \includegraphics[width=0.07\textwidth,height=0.07\textwidth]{figures/s12i7.png}& rank $45$
                                & \includegraphics[width=0.07\textwidth,height=0.07\textwidth]{figures/s13i2.png}& \includegraphics[width=0.07\textwidth,height=0.07\textwidth]{figures/13_4.png}& \includegraphics[width=0.07\textwidth,height=0.07\textwidth]{figures/59_4.png}& \includegraphics[width=0.07\textwidth,height=0.07\textwidth]{figures/s13i4.png}& rank $1$\\ \hline

\multirow{2}{*}{ITU-II}   &  \includegraphics[width=0.07\textwidth,height=0.07\textwidth]{figures/s3i10.png} & \includegraphics[width=0.07\textwidth,height=0.07\textwidth]{figures/149_09.png}& \includegraphics[width=0.07\textwidth,height=0.07\textwidth]{figures/138_9.png}& \includegraphics[width=0.07\textwidth,height=0.07\textwidth]{figures/s3i9.png}& rank $6$ 
								& \includegraphics[width=0.07\textwidth,height=0.07\textwidth]{figures/s6i1.png} & \includegraphics[width=0.07\textwidth,height=0.07\textwidth]{figures/18_7.png} & \includegraphics[width=0.07\textwidth,height=0.07\textwidth]{figures/14_7.png}& \includegraphics[width=0.07\textwidth,height=0.07\textwidth]{figures/s6i7.png}& rank $41$\\

& \includegraphics[width=0.07\textwidth,height=0.07\textwidth]{figures/s12i1.png}& \includegraphics[width=0.07\textwidth,height=0.07\textwidth]{figures/11_07.png} & \includegraphics[width=0.07\textwidth,height=0.07\textwidth]{figures/40_7.png}&  \includegraphics[width=0.07\textwidth,height=0.07\textwidth]{figures/s12i7.png}& rank $21$
                                & \includegraphics[width=0.07\textwidth,height=0.07\textwidth]{figures/s13i2.png}& \includegraphics[width=0.07\textwidth,height=0.07\textwidth]{figures/13_4.png}  & \includegraphics[width=0.07\textwidth,height=0.07\textwidth]{figures/59_4.png}& \includegraphics[width=0.07\textwidth,height=0.07\textwidth]{figures/s13i4.png}& rank $1$\\ 
                
\end{tabular}
}
\end{table*}

\textbf{Qualitative evaluation:} Last but not least we shows some qualitative examples of the recognition performance of all approaches participating in the UERC in Table~\ref{Tab: qualitative}. For this experiment we again use only images from the AWE dataset and select $1$ image per subject for the gallery, so the maximum rank of this experiment is $180$. In Table~\ref{Tab: qualitative} we show selected probe images and for each probe image, the best and second best match that was retrieved from the gallery. The correct galleries and information about the rank at which the correct gallery was retrieved is also given in the table. We can see that the UCCS approach selects the correct gallery for all of the selected examples despite different image characteristics. Other approaches make different types of errors, the most common being selecting a similarly looking image from another subject, but taken at same side of the head.

\end{document}